\pdfoutput=1
\documentclass[preprint,sort,compress,12pt]{elsarticle}
\usepackage{graphicx}
\usepackage{amsmath}
\usepackage{amssymb}
\usepackage{mathrsfs}
\usepackage{array}
\usepackage{upgreek}
\usepackage{float}
\usepackage{booktabs}
\usepackage{adjustbox}
\usepackage{makecell}
\usepackage[margin=1in]{geometry}
\usepackage[small,compact]{titlesec}
\usepackage[normal,bf,labelsep=colon,skip=2pt]{caption}
\usepackage[titles]{tocloft}
\usepackage{bold-extra}
\usepackage{hyperref}
\usepackage[ruled,vlined]{algorithm2e}
\usepackage{caption}
\usepackage{multirow}
\usepackage{lineno}
\usepackage{tabularx}
\usepackage{diagbox}
\usepackage{algorithmic}
\usepackage{subcaption}

\newcommand{\Tau}{\mathrm{T}}

\DeclareMathOperator*{\argmin}{arg\,min}

\usepackage{xcolor}
\hypersetup{
    colorlinks,
    linkcolor={red!50!black},
    citecolor={blue!50!black},
    urlcolor={blue!80!black}
}
\begin{document}
\begin{frontmatter}
\title{Stochastic and non-local closure modeling for nonlinear dynamical systems via latent score-based generative models}

\author{Xinghao Dong}
\author{Huchen Yang}
\author{Jin-Long Wu\corref{cor1}} \ead{jinlong.wu@wisc.edu; jinlong@caltech.edu} 
\cortext[cor1]{Corresponding author}

\address{Department of Mechanical Engineering, University of Wisconsin-Madison, Madison, WI, 53706, USA}

\begin{abstract}
    We propose a latent score-based generative AI framework for learning stochastic, non-local closure models and constitutive laws in nonlinear dynamical systems of computational mechanics. This work addresses a key challenge of modeling complex multiscale dynamical systems without a clear scale separation, for which numerically resolving all scales is prohibitively expensive, e.g., for engineering turbulent flows. While classical closure modeling methods leverage domain knowledge to approximate subgrid-scale phenomena, their deterministic and local assumptions can be too restrictive in regimes lacking a clear scale separation. Recent developments of diffusion-based stochastic models have shown promise in the context of closure modeling, but their prohibitive computational inference cost limits practical applications in many real-world settings. This work addresses this limitation by jointly training convolutional autoencoders with conditional diffusion models in latent space, significantly reducing the dimensionality of the sampling process while preserving essential physical characteristics. Numerical results demonstrate that the joint training approach helps discover a proper latent space that not only guarantees small reconstruction errors but also ensures good performance of the diffusion model in the latent space. When integrated into numerical simulations, the proposed stochastic modeling framework via latent conditional diffusion models achieves significant computational acceleration while maintaining comparable predictive accuracy to standard diffusion models in physical space.
\end{abstract}


\end{frontmatter}

\section{Introduction}
Complex nonlinear dynamical systems are fundamental in numerous scientific and engineering fields, including engineering turbulent flows~\cite{moin1997tackling,moin1998direct,duraisamy2019turbulence}, solid mechanics~\cite{holzapfel2002nonlinear,fuhg2024review},  geophysical flows and earth system modeling~\cite{majda2006nonlinear,schneider2017earth,donges2009complex,lai2024machine,schneider2024opinion}, and neuroscience for analyzing neural activity patterns \cite{jost2005dynamical,izhikevich2007dynamical}.  In these systems, explicitly resolving all relevant scales is often computationally prohibitive, e.g., it is often infeasible (and unnecessary for many cases) to resolve all the scales of a computational mechanics problem down to the micro-scales of molecular dynamics, which prompts the development of closure models and constitutive laws. Traditional closure models, e.g., in the context of turbulence, including Reynolds-averaged Navier–Stokes (RANS) models \cite{launder1983numerical, wilcox1998turbulence}, large eddy simulations (LES) \cite{smagorinsky1963general, deardorff1970numerical}, and moment closure techniques \cite{arnold1974stochastic, klimenko1999conditional}, often rely on deterministic parameterizations based on physical intuition, dimensional analysis, and a relatively small amount of data that facilitates the empirical model calibration. Moreover, these methods typically use local approximations and may struggle to capture regime transitions and emergent patterns inherent to complex systems. These limitations have recently motivated the development of more sophisticated modeling approaches, including stochastic and non-local modeling frameworks that better account for uncertainties and long-range interscale correlations \cite{pleim2007combined, chorin2015discrete, zhou2021learning, chen2023stochastic,wu2024learning,chen2025neural,sun2024lemon,you2024nonlocal,yu2024nonlocal,zhou2025neural,dong2025data,sanderse2025scientific}.

In recent decades, machine learning techniques have transformed various fields by uncovering complex patterns from large datasets without manually programmed rules \cite{lecun2015deep, goodfellow2016deep, cristianini2000introduction, kaelbling1996reinforcement}. These advancements have subsequently catalyzed the emergence of scientific machine learning (SciML), which applies these data-driven methodologies to scientific discovery and modeling \cite{baker2019workshop, bergen2019machine, carleo2019machine,wang2023scientific}. The particular strength of these approaches lies in their ability to capture hidden relationships, model nonlinear dependencies, and integrate diverse data sources. This, in turn, motivates the exploration of SciML approaches for representing the multiscale interactions of complex dynamical systems. For instance, deep learning techniques offer the flexibility to represent high-dimensional nonlinear functions, thereby enabling the encoding of sophisticated closure relationships via ML parameterizations \cite{maulik2019subgrid,rasp2018deep}. System identification methods have also been developed for the discovery of governing equations~\cite{rudy2017data, champion2019data,gao2025sparse} that facilitate interpretable data-driven models. On the other hand, researchers have also explored various ways to enhance the performance of data-driven models by integrating known physics, which not only improves interpretability but also strengthens generalizability, thus enabling more accurate predictions even in regimes with limited observational data \cite{wang2017physics,karpatne2017theory,wu2018physics,raissi2019physics,sun2020surrogate,wang2021physics,kashinath2021physics,karniadakis2021physics,willard2022integrating,baddoo2023physics,sharma2023review,yu2024learning}.

Among the development of SciML approaches, data-driven closure modeling~\cite{duraisamy2019turbulence,beck2019deep,wu2024learning} aims to build on existing physics-based models of complex dynamical systems and to identify a proper data-driven residual term that accounts for unresolved degrees of freedom. While many closure modeling approaches adopt the deterministic assumption and thus may fail to characterize uncertainties in complex systems, generative AI frameworks have become a promising alternative by viewing the closure term as a random object and characterizing it via probabilistic modeling methods such as \cite{rezende2015variational,goodfellow2016deep}. More recently, diffusion models such as denoising diffusion probabilistic models \cite{ho2020denoising} and score-based generative approaches \cite{song2019generative, song2020score} have emerged as recent advances of generative AI techniques and demonstrated strong performance on computer vision, audio synthesis, and text-to-image generation \cite{rombach2022high, kong2020diffwave, saharia2022photorealistic}. These methodologies operate by reversing a gradual noising process or through score-based sampling techniques, thereby offering several advantages over other generative model frameworks such as GANs and VAEs, e.g., notably more stable training dynamics, enhanced output diversity, and tractable likelihood estimation \cite{dhariwal2021diffusion, nichol2021improved}. 

These probabilistic frameworks have recently gained traction in SciML~\cite{stinis2019enforcing,gagne2020machine,yang2020physics,wu2020enforcing, yang2025active, yang2026bayesian, stinis2024sdyn}, with diffusion-based methods being adapted to model complex physical phenomena such as turbulent flows, weather patterns, and molecular dynamics \cite{fan2025neural, xu2022geodiff, song2021solving, li2024generative}. For example, Liu et al. \cite{liu2024confild} introduced ConFiLD as a conditional flow-matching diffusion model for generating diverse simulation data. Diffusion models have also been employed to reconstruct high-fidelity fields from low-fidelity approximations \cite{li2023multi, shu2023physics}. Unlike deterministic models that produce single-trajectory predictions, diffusion-based approaches offer several key advantages, such as accounting for rare yet significant extreme events, modeling the inherently stochastic behavior due to some unresolved scales, and providing more principled uncertainty quantification \cite{wen2023diffstg, rasul2021autoregressive, tashiro2021csdi}. The iterative reverse process in diffusion models also allows for incorporating physical constraints and conservation laws at each denoising step, ensuring that generated samples remain physically consistent while preserving the statistical characteristics observed in real systems \cite{shu2023physics, zhuang2025spatially, jacobsen2025cocogen}. In the context of stochastic and non-local closures, Dong et al. \cite{dong2025data, dong2026synergizing} demonstrated that combining neural operators \cite{li2020fourier} with diffusion models enables resolution-invariant stochastic closure modeling, allowing robust performance across different discretizations without retraining the model.

Despite their impressive generative capabilities, diffusion models face significant computational limitations during sampling. The sequential nature of the iterative denoising process requires hundreds to thousands of network evaluations, making standard diffusion models prohibitively expensive for applications like closure modeling, where evaluation may occur at every numerical time step during simulation. Various acceleration strategies have been proposed, including advanced numerical solvers \cite{lu2022dpm, liu2022pseudo}, knowledge distillation \cite{song2023consistency, salimans2022progressive, luhman2021knowledge}, and adaptive sampling schedules \cite{karras2022elucidating, kong2021fast}. On the other hand, latent diffusion models (LDMs) \cite{rombach2022high, vahdat2021score} address the computational challenge through a two-stage architecture: (i) an autoencoder first compresses high-dimensional data into a compact latent space, and (ii) a diffusion model operates exclusively within this reduced representation. Building upon this two-stage framework, Rombach et al. \cite{rombach2022high} implemented these principles in the stable diffusion architecture, demonstrating that this approach dramatically reduces computational requirements during both training and inference while preserving generative quality. This dimensionality reduction strategy yields multiple computational advantages, decreasing memory consumption and accelerating sampling by orders of magnitude \cite{esser2021taming, kingma2021variational}.

In this paper, we introduce a novel latent diffusion-based stochastic closure modeling framework for complex dynamical systems governed by partial differential equations. Our approach extends traditional conditional diffusion models by incorporating autoencoder-based dimensionality reduction that maps both the state variable and the unknown closure term into a lower-dimensional latent space. Within this latent space, a diffusion model is trained to learn the conditional probability distribution of the closure term given the corresponding state, thereby serving as an efficient, data-driven correction mechanism for classical physics-based solvers. Our primary contributions are as follows:

\begin{itemize}
    \item We leverage pretrained autoencoders to compress high-dimensional fields into a low-dimensional latent manifold, then train a latent diffusion model to perform conditional Langevin sampling there, delivering orders-of-magnitude faster inference. However, we show that the standard two-phase training pipeline compromises the accuracy of closure-term generation.
    
    \item We introduce an end-to-end training scheme that jointly optimizes the autoencoder and diffusion components, harmonizing reconstruction and score-matching objectives. This unified approach significantly improves generative accuracy while keeping the efficiency gains of latent-space sampling.
    
    \item We embed our latent diffusion closure into large-scale numerical simulations, enabling fast ensemble generation and uncertainty quantification. Our approach matches the simulation accuracy of full physical-space diffusion closures while delivering substantial acceleration, making it ideal for computationally demanding scientific applications.
\end{itemize}

\section{Methodology}\label{sec: methodology}
The governing equations for a general dynamical system studied in this work can be expressed in the following form:
\begin{equation}
    \label{eqn:true_system}
    \frac{\partial v}{\partial t} = \mathcal{M}(v),
\end{equation}
where the state $v\in\mathcal{V}$ and the nonlinear operator $\mathcal{M}:\mathcal{V}\to\mathcal{V}$ characterizes the dynamics of the state. Direct numerical simulation of this full system is often computationally infeasible, so we introduce a projection or filter $\mathcal{K}:\mathcal{V}\to\mathcal{W}$ to obtain a reduced state $V:=\mathcal{K}(v)\in\mathcal{W}$. The reduced dynamics are then approximated by
\begin{equation}
    \label{eqn:model_system}
    \frac{\partial V}{\partial t} = \overline{\mathcal{M}}(V),
\end{equation}
where $\overline{\mathcal{M}}:\mathcal{W}\to\mathcal{W}$ closes the system, often based on physical insight, but inevitably incurs modeling errors in many cases. For instance, most existing physics-based closure models are based on local and deterministic assumptions, which may not be valid for complex dynamical systems (e.g., turbulent flows) without a clear scale separation between the numerically resolved scales (i.e., the reduced-order state $V$) and the unresolved ones. In recent years, we have witnessed a rapid growth of high-fidelity simulation and experimental data. On the other hand, the advancement of machine learning and optimization techniques enables a systematic way to calibrate a complicated model with a large amount of data. Building upon this recent progress, we aim to go beyond the local and deterministic assumption of existing physics-based closure models by introducing a data-driven stochastic closure $U(V)$ to address the error of an existing closure model $\overline{\mathcal{M}}$:
\begin{equation}
    \label{eqn:corrected_system}
    \frac{\partial V}{\partial t} = \overline{\mathcal{M}}(V) + U(V),
\end{equation}
where $U: \mathcal{W} \rightarrow \mathcal{W}$ represents a machine-learning-based closure term that compensates for the approximation errors in $\overline{\mathcal{M}}(V)$, enabling more accurate representation of the true reduced-order dynamics. To achieve a stochastic formulation of the closure term, one could characterize $U$ as a stochastic field with non-Markovian dynamics, for example:
\begin{equation}
\label{eqn:closure_spde}
    \frac{\partial U}{\partial t} = h(U;\ V) + \xi,
\end{equation}
where $\xi$ represents space-time stochastic forcing and $h$ encompasses differential and/or integral operators governing the temporal evolution of $U$. However, explicitly learning or solving such an SPDE is computationally prohibitive. Therefore, this work adopts a different approach. We make a Markovian assumption—that the statistics of the closure $U$ at any time $t$ depend only on the resolved state $V(t)$. We thus aim to directly characterize the stationary, conditional probability distribution $p(U|V)$. As detailed in the following sections, we adopt a diffusion-based generative modeling approach to learn this time-independent conditional distribution from time-series data of $U$ and $V$. For the numerical simulation of the modeled system, the $U(V)$ term in Eq.~\eqref{eqn:corrected_system} is then evaluated by drawing a new sample $U \sim p(U|V)$ at each time step.

\begin{figure}[!tb]
    \centering
    \includegraphics[width=\linewidth]{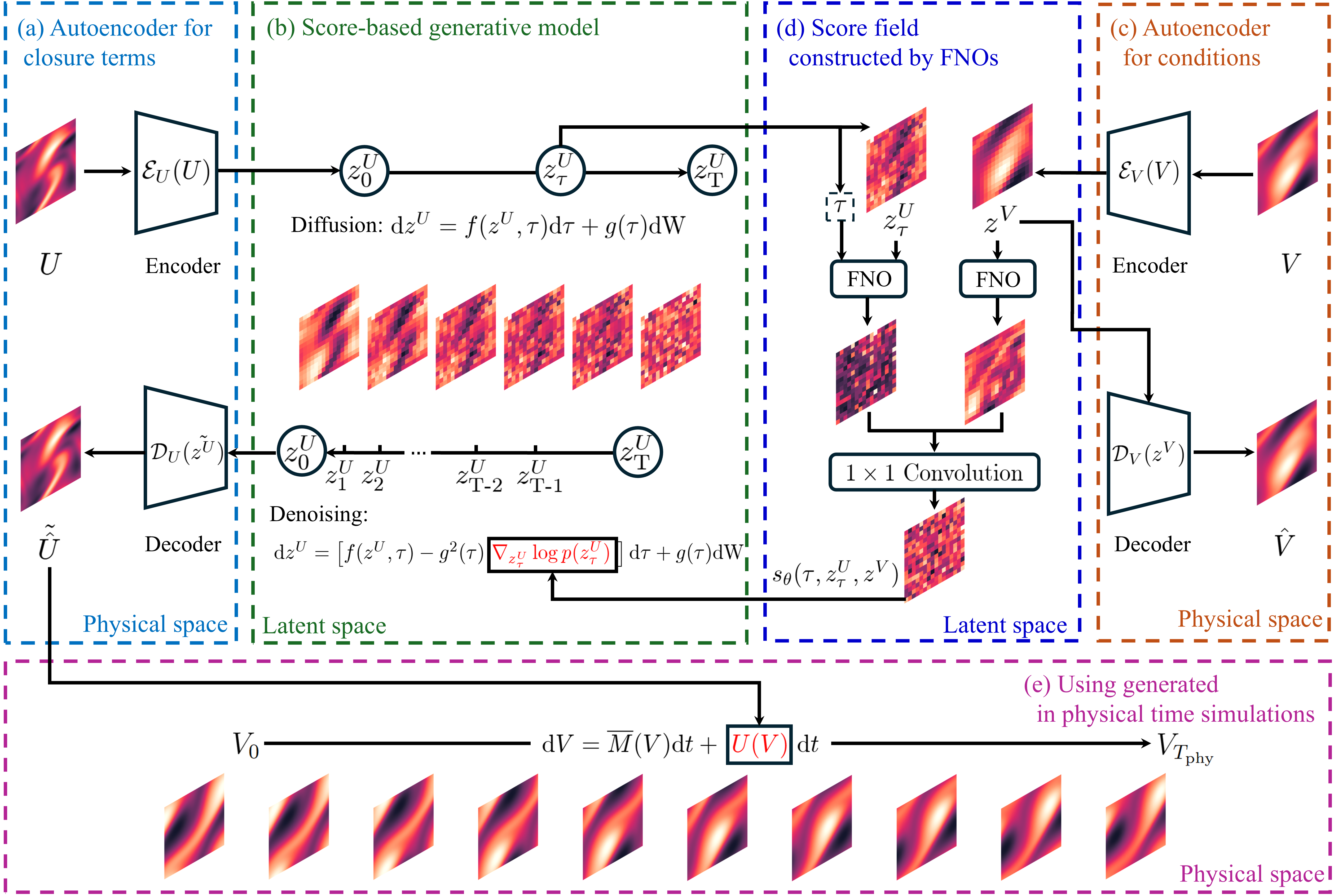}
    \caption{A schematic diagram of the proposed stochastic closure framework via latent conditional diffusion models. High-dimensional physical-space spatiotemporal fields (i.e., $U$ and $V$) are first encoded into low-dimensional latent spaces. A conditional diffusion model constructed via Fourier neural operators is then trained to capture the underlying probability distribution $p(z^U|z^V)$, where $z^U$ and $z^V$ denote the latent representations of the closure terms $U$ and the system states $V$, respectively. Latent samples of $z^U$ are then generated via conditioning on the corresponding latent representations of the system state, and the resulting samples are decoded back into the physical space to serve as a stochastic and non-local data-driven closure term for the numerical simulation of the system state $V$.}
    \label{fig: schematic}
\end{figure}

Diffusion-based models have recently been explored in~\cite{dong2025data, dong2026synergizing} as stochastic closure models, and it is worth noting that a key drawback of diffusion models lies in their iterative sampling procedure, which becomes computationally intensive when applied to problems with physical fields that demand a relatively high spatial and/or temporal resolution. Motivated by reducing the computational cost, here we develop a latent conditional diffusion model (L-CDM) that operates in a lower-dimensional latent space, offering a substantial reduction of computational cost for the reverse sampling process of a diffusion model. Numerical results confirm that our L-CDM architecture still efficiently captures the stochastic behavior of closure terms while dramatically reducing computational requirements compared to direct physical-space modeling. The complete framework is illustrated in Fig.~\ref{fig: schematic}, with detailed components and algorithms further described in Sections~\ref{ssec:latent_SBGM}--\ref{ssec: strategies}.

\subsection{Latent space formulation for score-based diffusion models}\label{ssec:latent_SBGM}
Dynamical systems governed by PDEs, such as turbulent flows, can feature a high-dimensional state space by discretizing the spatiotemporal fields. The high-dimensional state space poses a significant computational challenge for diffusion-based generative modeling, especially in the context of closure modeling, for which the sampling procedure of the diffusion model requires solving a reverse SDE at each time step when evaluating the closure term. To address this challenge, we introduce a latent space formulation that enables efficient learning and sampling while preserving essential physical characteristics.

\subsubsection{Dimensionality reduction via autoencoders}\label{sssec: AE}
We employ convolutional autoencoders to map high-dimensional physical fields to lower-dimensional latent representations:
\begin{align}\label{eqn:encoder_U}
z^U &= \mathcal{E}_U(U), \quad U \in \mathbb{R}^{d_U} \rightarrow z^U \in \mathbb{R}^{l_U}, \\
\label{eqn:encoder_V}
z^V &= \mathcal{E}_V(V), \quad V \in \mathbb{R}^{d_V} \rightarrow z^V \in \mathbb{R}^{l_V},
\end{align}
where $l_U \ll d_U$ and $l_V \ll d_V$. For 2D spatial fields, the input dimensions can be expressed as $d_U = H \times W \times C_U$ and $d_V = H \times W \times C_V$, with $H \times W$ representing the spatial resolution and $C_U, C_V$ the number of channels. The corresponding decoder reconstructs the physical fields from latent representations:
\begin{equation}\label{eqn:decoder}
\hat{U} = \mathcal{D}_U(z^U),\quad \hat{V} = \mathcal{D}_V(z^V).
\end{equation}

The autoencoder parameters are optimized by minimizing a reconstruction loss:
\begin{align}\label{eqn:ae_loss}
\mathcal{L}^U_{\text{AE}} &= \mathbb{E}_{U \sim p(U)} \| U - \mathcal{D}_U(\mathcal{E}_U(U)) \|_2^2, \\
\mathcal{L}^V_{\text{AE}} &= \mathbb{E}_{V \sim p(V)} \| V - \mathcal{D}_V(\mathcal{E}_V(V)) \|_2^2.
\end{align}

These encoders and decoders are designed to preserve essential multiscale structures while eliminating redundant information, creating a compact representation amenable to diffusion modeling. In this work, convolutional layers with residual connections are employed to capture multi-scale spatial correlations critical in turbulent systems, effectively transforming high-dimensional fields to compact latent representations. This preservation of multiscale structures is achieved implicitly by the hierarchical nature of this convolutional architecture, for which more details are presented in \ref{sec:training_details}.

The selection of these latent dimensions ($l_U$ and $l_V$) is treated as hyperparameters governed by a fundamental trade-off. The dimension must be large enough to encode essential multiscale physical characteristics and to ensure high reconstruction fidelity. Conversely, it needs to be as small as possible to reduce the computational cost of drawing samples in the latent space via the trained diffusion model. In this work, the number of latent dimensions was determined empirically by gradually reducing it before observing a significant drop in reconstruction performance.

\subsubsection{Unconditional score-based models in latent space}\label{sssec: DM}
Once the dimensionality reduction framework is established, we can formulate a score-based generative model in the latent space. Instead of directly approximating the probability distribution function $p(z^U)$, we learn the score function $\nabla_{z^U} \log p(z^U)$, defined as the gradient of the log probability density. The score function does not require a tractable normalizing constant and can be trained directly from data samples via score matching techniques.

To address challenges of inaccurate score estimation in regions of low data density, we employ a continuous-time noise perturbation process defined by a stochastic differential equation (SDE). For our latent space formulation, we adopt a variance exploding (VE) SDE \cite{song2020score}:
\begin{equation}\label{eqn:latent_forwardSDE}
\mathrm{d}z^U_\tau = \sigma^\tau\mathrm{dW},
\end{equation}
where $\mathrm{W}$ denotes a standard Wiener process, $\tau \in [0,T]$ represents diffusion time, $\sigma > 0$ is a scaling constant, and $\sigma^\tau$ denotes $\sigma$ raised to the power of $\tau$. This formulation yields a Gaussian transition kernel:
\begin{equation}\label{eqn:latent_transitionKernel}
p(z^U_\tau \mid z^U_0) = \mathcal{N}\left(z^U_0,\, \Sigma(\tau)\right), \quad \text{with} \quad \Sigma(\tau)=\frac{1}{2\log\sigma}\left(\sigma^{2\tau}-1\right) I.
\end{equation}

At $\tau=0$, the distribution $p(z_{0}^{U})$ corresponds to the true latent distribution of our data. The final time $T$ is chosen as 1 following the standard convention~\cite{song2020score}. With a sufficiently large $\sigma$ (selected as $\sigma=25$ in this work), the accumulated noise variance dominates the data signal at $\tau=T$, ensuring that the distribution $p(z_{T}^{U})$ approaches a tractable Gaussian prior:
\begin{equation}\label{eqn:latent_prior}
p(z^U_T) = \int p(z^U_0) \cdot \mathcal{N}\left(z^U_0, \frac{1}{2 \log \sigma}\left(\sigma^{2T}-1\right) I\right) \mathrm{d}z^U_0 \approx \mathcal{N}\left(0, \frac{1}{2 \log \sigma}\left(\sigma^{2T}-1\right) I\right).
\end{equation}

To train the score-based model in latent space, we leverage the equivalence between explicit score matching (ESM) and denoising score matching (DSM):
\begin{equation}\label{eqn:latent_ESMDSM}
\mathbb{E}_{z^U_\tau \sim p(z^U_\tau)} \left \| \nabla_{z^U_\tau} \log p(z^U_\tau) - s_\theta \right \|^2_2 = \mathbb{E}_{z^U_\tau \sim p(z^U_\tau \mid z^U_0)}\mathbb{E}_{z^U_0 \sim p(z^U_0)} \left \| \nabla_{z^U_\tau} \log p(z^U_\tau \mid z^U_0) - s_\theta\right \|^2_2 + C,
\end{equation}
where $s_\theta$ represents the model to be trained and $C$ is a constant independent of $\theta$. Detailed derivations of this equivalence can be found in \ref{sec:score_matching_proof}. Since the true score function required by ESM is typically inaccessible, DSM provides a practical training objective:
\begin{equation}\label{eqn:latent_unconscorematching}
\theta^* = \argmin_\theta \left \{ \mathbb{E}_{\tau \sim \mathcal{U}(0, T)} \mathbb{E}_{z^U_\tau \sim p(z^U_\tau \mid z^U_0)} \mathbb{E}_{z^U_0 \sim p(z^U_0)} \left \| \nabla_{z^U_\tau} \log p(z^U_\tau \mid z^U_0) - s_\theta(\tau, z^U_\tau)\right \|^2_2\right \}.
\end{equation}

An advantage of this formulation is that $\nabla_{z^U_\tau} \log p(z^U_\tau \mid z^U_0)$ has a closed-form expression due to the Gaussian transition kernel:
\begin{equation}\label{eqn:latent_closedformgrad}
\nabla_{z^U_\tau} \log p(z^U_\tau \mid z^U_0) = -\Sigma(\tau)^{-1}\left(z^U_\tau - z^U_0\right).
\end{equation}

\subsubsection{Conditional latent diffusion for closure modeling}\label{sssec: CDM}

In this work, we focus on stochastic closure modeling, for which the samples of the closure term $U$ are obtained from a conditional probability distribution with respect to the system state $V$. In the context of latent diffusion models, we work with the latent representations $z^U$ and $z^V$ instead and employ a conditional latent diffusion model to characterize the distribution $p(z^U \mid z^V)$.

The forward diffusion process remains identical to the unconditional case, as the noise perturbation acts independently of the conditioning variable:
\begin{equation}\label{eqn:latent_conditionalforward}
p(z^U_\tau \mid z^U_0, z^V) = p(z^U_\tau \mid z^U_0).
\end{equation}

On the other hand, the training objective must incorporate the conditioning variable $z^V$. More specifically, the training objective function is formulated as:
\begin{equation}\label{eqn:latent_conditionscorematching}
\theta^* = \argmin_\theta \left \{ \mathbb{E}_{\tau \sim \mathcal{U}(0, T)} \mathbb{E}_{z^U_\tau \sim p(z^U_\tau \mid z^U_0)} \mathbb{E}_{z^U_0, z^V \sim p(z^U_0, z^V)} \left \| \nabla_{z^U_\tau} \log p(z^U_\tau \mid z^U_0) - s_\theta(\tau, z^U_\tau, z^V)\right \|^2_2\right \}.
\end{equation}

Note that we take the expectation over the joint distribution $p(z^U_0, z^V)$ rather than the conditional distribution $p(z^U_0 \mid z^V)$, as the latter would be difficult to sample from directly. This formulation allows us to train the score model using paired samples from our dataset, with $z^V$ serving as a conditioning input to the score-based model.

Once the score function $s_\theta(\tau, z^U_\tau, z^V)$ is trained, we can generate samples from the target distribution $p(z^U \mid z^V)$ by solving the reverse SDE:
\begin{equation}\label{eqn:latent_reverseSDE}
\mathrm{d}z^U_\tau = -\sigma^{2\tau}\nabla_{z^U_\tau}\log p(z^U_\tau\mid z^V)\mathrm{d}\tau + \sigma^\tau\mathrm{d}\bar{\mathrm{W}},
\end{equation}
where $\bar{\mathrm{W}}$ is a reverse-time Wiener process. In implementation, we replace the true score with our learned approximation $s_\theta(\tau, z^U_\tau, z^V)$ and integrate numerically from $\tau=T$ to $\tau=0$. For numerical integration, the Euler-Maruyama scheme is employed:
\begin{equation}\label{eqn:latent_eulermaruyama}
z^U_{\tau - \Delta \tau} = z^U_\tau + \sigma^{2\tau}s_\theta(\tau, z^U_\tau, z^V)\Delta\tau + \sigma^\tau\sqrt{\Delta\tau}z,
\end{equation}
where $z \sim \mathcal{N}(0, I)$ and $\Delta\tau$ represents the time step. Starting from a sample $z^U_T \sim \mathcal{N}(0, \Sigma(T))$, this process gradually transforms Gaussian noise into a sample from the conditional distribution $p(z^U_0 \mid z^V)$. With the sampled $z^U$ from the diffusion model, the final step involves decoding the sample in latent space back to physical space, i.e., $U = \mathcal{D}_U(z^U_0)$.

In~\cite{dong2025data}, an adaptive time scheme was introduced to enhance the computational efficiency of the sampling procedure in the conditional diffusion model, to reduce the computational cost of the stochastic closure model. Here, we adopt a similar strategy of an adaptive scheme to ensure an efficient sampling procedure of the proposed latent diffusion model. More specifically, standard diffusion models typically employ uniform time discretization during reverse sampling, allocating computational resources inefficiently. A non-uniform time-stepping scheme was proposed in \cite{karras2022elucidating}:
\begin{equation}\label{eqn: adaptive_scheme}
    \tau_i = \left(\tau_{\max}^{1/\rho} + \frac{i}{N}\Bigl(\tau_{\min}^{1/\rho} - \tau_{\max}^{1/\rho}\Bigr)\right)^\rho, \quad i=0,1,\ldots,N,
\end{equation}
where $\rho=7$ controls the distribution of steps, $\tau_{\max}=T$ represents the maximum diffusion time, $\tau_{\min}=10^{-3}$ ensures numerical stability, and $N$ denotes the total number of reverse-diffusion steps. This formulation allocates larger steps when noise levels are high (early in the reverse process) and progressively smaller steps as $\tau$ approaches zero. 

The complete sampling process, which implements the adaptive Euler-Maruyama integration described above and corresponds to the generation and decoding steps illustrated in Fig.~\ref{fig: schematic}, is summarized in Algorithm~\ref{alg:sampling}.

\begin{algorithm}[tbp]
    \caption{Sampling using latent conditional diffusion models}
    \begin{algorithmic}[1]
        \STATE \textbf{Input:} condition $V$, encoder $\phi_V$, decoder $\phi_U$, score $s_\theta$, number of steps $N$
        \STATE $z^V \leftarrow E_{\phi_V}(V)$
        \STATE $Z^U_{\Tau}\sim\mathcal{N}\bigl(0,\tfrac{\sigma^{2\Tau}-1}{2\ln\sigma}I\bigr)$ 
        \STATE Set $\tau_{\max}=\Tau$, $\tau_{\min}=10^{-3}$, $\rho=7$
        \FOR{$i=0$ \TO\ $N$}
            \STATE $\tau_i = \bigl(\tau_{\max}^{1/\rho} + \tfrac{i}{N}(\tau_{\min}^{1/\rho}-\tau_{\max}^{1/\rho})\bigr)^{\rho}$
        \ENDFOR
        \FOR{$i=0$ \TO\ $N-1$}
            \STATE $\Delta\tau \leftarrow \tau_i - \tau_{i+1}$
            \STATE Sample $\epsilon\sim\mathcal{N}(0,I)$
            \STATE $Z^U_{\tau_{i+1}} \leftarrow Z^U_{\tau_i}
                + \sigma^{2\tau_i}\,s_\theta(\tau_i,Z^U_{\tau_i},z^V)\,\Delta\tau
                + \sigma^{\tau_i}\sqrt{\Delta\tau}\,\epsilon$ 
        \ENDFOR
        \STATE $U \leftarrow D_{\phi_U}(Z^U_{\tau_N})$ 
    \end{algorithmic}
    \label{alg:sampling}
\end{algorithm}

\subsection{Modeling non-local spatial dependencies}
A primary limitation of classical closure models is their local assumption—that the closure $U$ at a point $\mathrm{x}$ depends only on the resolved state $V$ or other related physical variables at that same point. This fails to capture the long-range spatial correlations inherent in many complex systems, such as those from pressure fields or turbulent backscatter, which are known to be non-local phenomena~\cite{duraisamy2019turbulence, pleim2007combined}. To empirically illustrate the necessity of non-local modeling for the numerical example in this work, we also perform an ablation study (see~\ref{sec: ablation_nonlocal}), which confirms that local-in-space models are fundamentally insufficient. On the other hand, our framework is non-local by design to address this critical gap~\cite{zhou2021learning, dong2025data}.

The non-locality is enabled by the Fourier Neural Operator (FNO)~\cite{li2020fourier} used to construct the score function $s_\theta$. As shown in Fig.~\ref{fig: schematic}(d), FNOs are used to construct the conditional score-based model in the latent space. By design, the FNO performs its critical operation—a learned convolution $K$—in the Fourier domain, as defined by the Convolution Theorem:
\begin{equation}
    (K\mathcal{Q})(\mathbf{x}) = \mathcal{F}^{-1}\left(R \cdot (\mathcal{F}\mathcal{Q})\right)(\mathbf{x})
\end{equation}
where $\mathcal{F}$ is the Fourier transform, $\mathcal{F}^{-1}$ is its inverse, $\mathcal{Q}$ is the input field, and $R$ is the learned parameter tensor. This multiplication in the spectral domain is equivalent to a global convolution in the physical domain. This intrinsic property means that the score estimated at any single point $\mathbf{x}$ is a function of the input fields across the entire domain, making the FNO an ideal choice for learning global physical dependencies.

The autoencoders also provide two additional non-local mechanisms. First, they are built as deep convolutional neural networks (CNNs)~\cite{krizhevsky2012imagenet}. Through a hierarchy of convolutional and strided downsampling layers, a global receptive field is achieved: by the time the input field is encoded into the latent bottleneck, each latent neuron is influenced by the entire global input field. Second, we explicitly include self-attention modules within the autoencoder's bottleneck, which directly compute the interaction between all spatial patches~\cite{vaswani2017attention}. This is achieved by computing a weighted sum of all input features (values $\mathcal{V}$), where the weights are derived from the similarity (queries $Q$ and keys $K$) between all pairs of features, as defined by the canonical scaled dot-product attention:
\begin{equation}
    \text{Attention}(Q, K, \mathcal{V}) = \text{softmax}\left(\frac{QK^T}{\sqrt{d_k}}\right)\mathcal{V}.
\end{equation}
This provides a second, explicit mechanism for modeling non-local dependencies, similar in principle to other nonlocal attention operators.

This combination—a global FNO for the score model and an autoencoder employing both a deep hierarchical architecture and explicit self-attention—makes our framework fully non-local. Many data-driven closures use standard CNNs, where the output at any point is only influenced by a small local neighborhood of the input. The novelty of our design is the synthesis of these inherently non-local components within a stochastic and computationally efficient latent-space framework, enabling it to capture the complex, global dependencies of the target physical system.

\subsection{Training strategies for latent conditional diffusion models}\label{ssec: strategies}

The latent diffusion model hinges critically on the training procedure that balances the representation power of the latent space and the performance of the diffusion model in that latent space. While conventional LDM implementations \cite{rombach2022high, vahdat2021score, liu2023audioldm, takagi2023high, podell2024sdxl} predominantly employ a two-phase training procedure, i.e., autoencoders and diffusion models are trained separately, our investigation reveals fundamental limitations of this strategy for closure modeling applications. This section presents both the standard two-phase training framework and our proposed joint optimization strategy, highlighting key innovations for improved performance.

\subsubsection{Conventional two-phase training}\label{sssec: decoupled}
The established convention for latent diffusion modeling \cite{rombach2022high} implements a two-phase training pipeline. First, autoencoder networks are optimized independently to minimize reconstruction error:

\begin{equation}\label{eqn: AE_loss_full}
    \mathcal{L}_{\text{AE}} = \frac{1}{N}\sum_{i=1}^{N} \left( \|U_i - \mathcal{D}_U(\mathcal{E}_U(U_i))\|_2^2 + \|V_i - \mathcal{D}_{V}(\mathcal{E}_{V}(V_i))\|_2^2 \right),
\end{equation}

where $N$ denotes the total number of training samples. Secondly, these autoencoders are fixed, and the diffusion model is trained in the corresponding latent space, using the score matching objective from Eq.~\eqref{eqn:latent_conditionscorematching}.

This two-phase training approach offers several practical advantages: (i) it simplifies the training pipeline, provides more stable optimization dynamics, and (ii) it allows for reusing the same pretrained autoencoders across multiple generative tasks. However, it suffers from a fundamental limitation: the latent space is optimized solely for the performance of reconstruction, without considering the subsequent generative process. Therefore, the target conditional distribution in the latent space may be difficult to capture via diffusion models and thus leads to unsatisfactory overall performance, especially when both the target and conditioning variables are continuous objects. In the context of stochastic closure modeling, where capturing the conditional relationship between state and closure term is crucial, the limitation of the conventional two-phase training procedure can significantly impact performance.

\subsubsection{Joint training with latent space regularization}\label{sssec: joint}
To address these limitations, we introduce a joint optimization framework where the autoencoder and the diffusion model in latent space are trained simultaneously. This approach allows the adaptive tuning of latent representation to accommodate the performance of the diffusion model. In general, our joint training combines three complementary objectives:

\begin{equation}\label{eqn: joint_loss_full}
    \mathcal{L}_{\text{joint}} = \mathcal{L}_{\text{AE}} + \lambda_{\text{score}}\, \mathcal{L}_{\text{score}} + \lambda_{\text{KL}}\, \mathcal{L}_{\text{KL}},
\end{equation}
where $\mathcal{L}_{\text{AE}}$ denotes the autoencoder reconstruction loss, $\mathcal{L}_{\text{score}}$ corresponds to the score-based diffusion model loss, and $\mathcal{L}_{\text{KL}}$ the regularization of latent space to mitigate the collapsing of the latent representations. More specifically, the reconstruction loss ensures faithful representation of both fields while emphasizing the closure term:

\begin{equation}\label{eq:AE_loss_joint_modified}
    \mathcal{L}_{\text{AE}} = \frac{1}{N}\sum_{i=1}^{N} \left( \lambda_U\, \|U_i - \mathcal{D}_U(\mathcal{E}_U(U_i))\|_2^2 + \lambda_{V}\, \|V_i - \mathcal{D}_{V}(\mathcal{E}_{V}(V_i))\|_2^2 \right),
\end{equation}

where typically $\lambda_U > \lambda_{V}$ to prioritize accurate reconstruction of the closure term $U$. On the other hand, the score matching component 
$\mathcal{L}_{\text{score}}$ is the denoising score matching loss explicitly defined in Eq.~\eqref{eqn:latent_conditionscorematching}, training the diffusion model jointly with the latent space.

A critical challenge in joint training, identified in recent literature \cite{li2024study, nguyen2025improving, dieleman2022continuous, gao2024empowering}, is known as latent space collapse, i.e., a phenomenon where the encoder maps diverse inputs to deterministic, low-variance latent representations. This latent space collapse occurs because the diffusion model can trivially minimize its objective when the latent distribution becomes degenerate, creating a perverse incentive that undermines stochasticity essential for generative modeling.

To counteract the potential latent space collapse, we employ Kullback-Leibler divergence to formulate a regularization loss term:

\begin{equation}\label{eq:KL_loss_joint}
    \mathcal{L}_{\text{KL}} = \text{KL}\Bigl(q(z^U) \,\|\, p(z^U)\Bigr),
\end{equation}

where $q(z^U)$ represents the empirical distribution of encoded vectors and $p(z^U)$ is a prior distribution, typically chosen as a standard Gaussian. 
This $\mathcal{L}_\text{KL}$ term is implemented as a batch-level regularizer on the empirical distribution of latent vectors. For a given training batch, we first compute the mean ($\mu_\text{batch}$) and variance ($\sigma^2_\text{batch}$) of the latent vectors across the batch dimension. These empirical statistics are then used in the analytical KL-divergence formula, $\text{KL}(\mathcal{N}(\mu_\text{batch}, \sigma^2_\text{batch} I) || \mathcal{N}(0, I))$, as $0.5 \sum (\sigma^2_\text{batch} + \mu_\text{batch}^2 - 1 - \log(\sigma^2_\text{batch}))$. This loss penalizes the entire batch's distribution for deviating from a standard Gaussian, which we found to be a robust method for preventing the latent space collapse associated with joint training. This regularization enforces distributional properties in the latent space thereby maintaining sufficient variability of the latent representation and thus enhancing the learning performance of a diffusion model in the latent space. While latent collapse remains an active research challenge~\cite{nguyen2025improving, dieleman2022continuous}, our empirical results demonstrate that careful calibration of regularization strengths significantly enhances stability and generative quality.

It is worth noting that the loss function $\mathcal{L}_\text{joint}$ of the joint training is snapshot-based, meaning it is computed on individual, statistically independent pairs of $(U, V)$ sampled from the training dataset. This is consistent with our goal of learning a modular, Markovian closure term $p(U|V)$ that acts as an instantaneous statistical correction. This approach differs from end-to-end, time-unrolled training strategies (e.g., differentiable solvers), which are designed for direct temporal forecasting rather than creating a modular closure. Thus, "joint training" in this work refers to the simultaneous optimization of the model components (the autoencoders and the diffusion model), not an integration over multiple time steps of a simulation. While this creates a more complex optimization landscape, this joint training strategy does not increase the model's architectural complexity; the total number of parameters is identical to the two-phase LDM, as the same components are simply trained in a coupled manner. The goal is to guide the optimization to a solution that supports both reconstruction fidelity and a well-structured latent space. The hyperparameters $\lambda_U$, $\lambda_{V}$, $\lambda_{\text{score}}$ and $\lambda_{\text{KL}}$ that balance these objectives are determined through systematic validation experiments, with detailed configurations provided in~\ref{sec:training_details}. The full joint optimization procedure, which simultaneously trains the autoencoder components and the latent score-based model according to the composite loss in Eq.~\eqref{eqn: joint_loss_full}, is detailed in Algorithm~\ref{alg:training}.

This joint optimization approach represents a key methodological contribution for stochastic closure modeling, producing latent representations simultaneously optimized for reconstruction fidelity and conditional generation capabilities. By aligning the latent structure with both objectives, we achieve significant improvements in capturing the complex statistical relationships between resolved states and closure terms, which is essential for the stochastic reduced-order modeling of complex dynamical systems without a clear scale separation, e.g., subgrid-scale physics in turbulent flows or cloud dynamics.

\begin{algorithm}[!htbp]
    \caption{Joint training with latent space regularization}
    \begin{algorithmic}[1]
        \STATE \textbf{Input:} encoder/decoder params $\phi_U,\phi_V$, score params $\theta$; weights $\lambda_U,\lambda_V,\lambda_{\text{KL}}$
        \REPEAT
            \STATE Sample a mini‐batch $(U,V)\sim p(U_0,V)$ 
            \STATE $z^U_0 \leftarrow E_{\phi_U}(U)$, \quad $z^V \leftarrow E_{\phi_V}(V)$ 
            \STATE Sample $\tau\sim\mathcal{U}(0,\Tau)$, \quad $\epsilon\sim\mathcal{N}(0,I)$ 
            \STATE $z^U_{\tau} \leftarrow z^U_0 + \sqrt{\frac{\sigma^{2\tau}-1}{2\ln\sigma}}\,\epsilon$ 
            \STATE $\mathcal{L}_{\text{score}} \leftarrow \bigl\|s_\theta(\tau,z^U_{\tau},z^V)
                + \tfrac{\sqrt{2\ln\sigma}}{\sqrt{\sigma^{2\tau}-1}}\,\epsilon\bigr\|^2$ 
            \STATE $\mathcal{L}_{\text{rec}}^U \leftarrow \|D_{\phi_U}(z^U_0)-U\|^2$ \quad
                   $\mathcal{L}_{\text{rec}}^V \leftarrow \|D_{\phi_V}(z^V)-V\|^2$ 
            \STATE $\mathcal{L}_{\text{KL}} \leftarrow \mathrm{KL}\bigl(q_\text{batch}(z^U_0)\,\|\,
                   \mathcal{N}(0,I)\bigr)$ 
            \STATE $\mathcal{L} \leftarrow \lambda_{\text{score}}\,\mathcal{L}_{\text{score}}
                   + \lambda_U\,\mathcal{L}_{\text{rec}}^U
                   + \lambda_V\,\mathcal{L}_{\text{rec}}^V
                   + \lambda_{\text{KL}}\,\mathcal{L}_{\text{KL}}$
            \STATE Update $\theta,\phi_U,\phi_V$ via gradients of $\mathcal{L}$
        \UNTIL{convergence}
    \end{algorithmic}
    \label{alg:training}
\end{algorithm}

\section{Numerical results}\label{sec: numerical_results}

In our numerical experiments, we evaluate the performance of the proposed latent diffusion-based closure modeling framework on a two-dimensional Kolmogorov flow. More specifically, we consider the following stochastic 2-D Navier--Stokes equation for a viscous, incompressible fluid in vorticity form on the unit torus:
\begin{equation}\label{2dNS}
    \begin{aligned}
        \frac{\partial \omega(\mathrm{x}, t)}{\partial t} = -u(\mathrm{x}, t) \cdot \nabla\omega(\mathrm{x}, t) &+ f(\mathrm{x}) + \nu\nabla^2\omega(\mathrm{x}, t) + \beta \xi, \quad &&(t, \mathrm{x}) \in (0, T_{\text{phy}}] \times (0, L)^2\\
        \nabla \cdot u(\mathrm{x}, t) &= 0, &&(t, \mathrm{x}) \in (0, T_{\text{phy}}] \times (0, L)^2\\
        \omega(\mathrm{x}, 0) &= \omega_0(\mathrm{x}), &&\mathrm{x} \in (0, L)^2
    \end{aligned}
\end{equation}

This system serves as a prototypical representation of turbulent flows in geophysical, environmental, and engineering applications. Here, $u$ denotes the divergence-free velocity field, and $\omega = \nabla \times u$ represents the corresponding vorticity. We set the viscosity coefficient $\nu = 10^{-3}$ and impose periodic boundary conditions on a square domain of length $L=1$. The initial vorticity is generated from a Gaussian random field $\omega_0 \sim \mathcal{N}\left(0,\, 7^{3/2}(-\Delta + 49I)^{-5/2}\right)$, ensuring a broad spectrum of scales in the initial state. The deterministic forcing function $f(\mathrm{x}) = 0.1\left(\sin(2\pi(x+y)) + \cos(2\pi(x+y))\right)$ serves as a spatially varying driver that, together with the periodic boundaries, promotes the formation of coherent, large-scale structures within the flow field. Additionally, we incorporate a small-amplitude stochastic forcing term $\xi = \mathrm{d}\mathrm{W}/\mathrm{d}t$ scaled by $\beta = 5 \times 10^{-5}$, which mimics the effects of unresolved small-scale dynamics. In this flow regime, the combination of low viscosity and moderate forcing generates turbulent flow where nonlinear convection drives chaotic energy transfers and vortex interactions, capturing the complex interplay between local dissipation and non-local energy transport.

In this work, we assume that the right-hand side of Eq.~\eqref{2dNS} is only partially known. Specifically, the stochastic nonlinear convection term 
\begin{equation}\label{eqn: convection}
    H(\mathrm{x}, t) = -u(\mathrm{x}, t) \cdot \nabla\omega(\mathrm{x}, t) + \beta \xi
\end{equation}
is assumed to be unknown. We treat this term as our target for closure modeling because it is inherently stochastic and exhibits non-local dependencies on the resolved system state $\omega(\mathrm{x}, t)$, making it a suitable canonical example to demonstrate the capabilities of our methodology. We emphasize that this test case is a controlled toy problem with partially unknown physics: rather than deriving $H$ from an explicit spatial filter, we remove the full stochastic nonlinear convection term in Eq.~\eqref{eqn: convection} from the simulator and treat it as unknown. Since $H$ explicitly contains the intrinsic stochastic forcing component $\beta\xi$, recovering $H$ requires learning a conditional distribution $p(H\mid\omega)$, rather than only the conditional mean $\mathbb{E}[H\mid\omega]$.

We solve Eq.~\eqref{2dNS} numerically using a pseudo-spectral method detailed in \ref{ssec: pseudospectral}, with time integration performed via the Crank-Nicolson scheme described in \ref{ssec: cranknicolson}. Simulations are conducted on a uniform $256 \times 256$ grid with a fixed time step $\Delta t = 10^{-3}$. During data generation, we record the system state and corresponding convection term every 100 steps—equivalent to a physical interval of $0.1\ \mathrm{s}$—and uniformly subsample all fields (including the resolved state $\omega$ and closure term $H$) to a $64\times64$ resolution for both the training and test sets. We simulate 100 independent time series, each spanning 40 seconds, but utilize only the data from 20 to 40 seconds, with the goal of allowing the system to reach statistical equilibrium during the initial 20 seconds. Data from 90 time series are used for training, with the remaining 10 reserved for testing.

To quantitatively evaluate the performance of our modeling framework, we employ two metrics: the mean squared error (MSE) and the relative Frobenius norm error, defined respectively as:
\begin{align}
    D_{\text{MSE}}(H, \hat{H}) &= \frac{\|H - \hat{H}\|_F^2}{N},\\[2mm]
    D_{\text{RE}}(H, \hat{H}) &= \frac{\|H - \hat{H}\|_F}{\|H\|_F},
\end{align}
where $H$ denotes the ground truth field, $\hat{H}$ represents the model-generated field, $\|\cdot\|_F$ indicates the Frobenius norm, and $N$ is the total number of grid points in the spatial domain.

For the remainder of this section, the experimental results are organized to highlight three aspects of this work:

\begin{itemize}
    \item The conventional two-phase training of the latent diffusion framework successfully compresses $64 \times 64$ physical fields into $16 \times 16$ latent representations. Despite the computational gains via working with a lower-dimensional latent space, the performance of the latent diffusion model is noticeably worse than directly training a diffusion model in physical space. Detailed results can be found in Section~\ref{ssec: decoupled}.
    
    \item The proposed joint optimization of autoencoder and diffusion components significantly enhances the performance of the generative model relative to the conventional two-phase training approach. Detailed results can be found in Section~\ref{ssec: joint_results}.
    
    \item When deployed within numerical simulations, the stochastic closure model enabled by the latent space diffusion framework achieves much faster ensemble simulations (e.g., $\sim 10\times$ acceleration for the 2-D Kolmogorov flow), compared to the one that exploits the standard diffusion framework in the physical space. This enhanced computational efficiency enables practical uncertainty quantification studies that would otherwise be computationally prohibitive. Detailed results can be found in Section~\ref{sec: surrogate}.
\end{itemize}

\subsection{Diffusion model in a separately trained latent space}\label{ssec: decoupled}

We first study the computational efficiency by comparing latent diffusion models against a physical-space baseline. The latent space is trained separately, and the main purpose is to demonstrate the reduction of computational cost due to dimensional reduction, while illustrating the potential limitation of learning the latent diffusion model with a separately trained latent space.

As a reference point, we implement a physical-space conditional diffusion model (P-CDM) following the methodology established in \cite{dong2025data}, where diffusion models generate closure terms $\hat{H}(\mathrm{x},t)$ conditioned on the resolved state $\omega(\mathrm{x},t)$ at the original $64 \times 64$ resolution.

\begin{figure}[!tb]
    \centering
    \includegraphics[width=\linewidth]{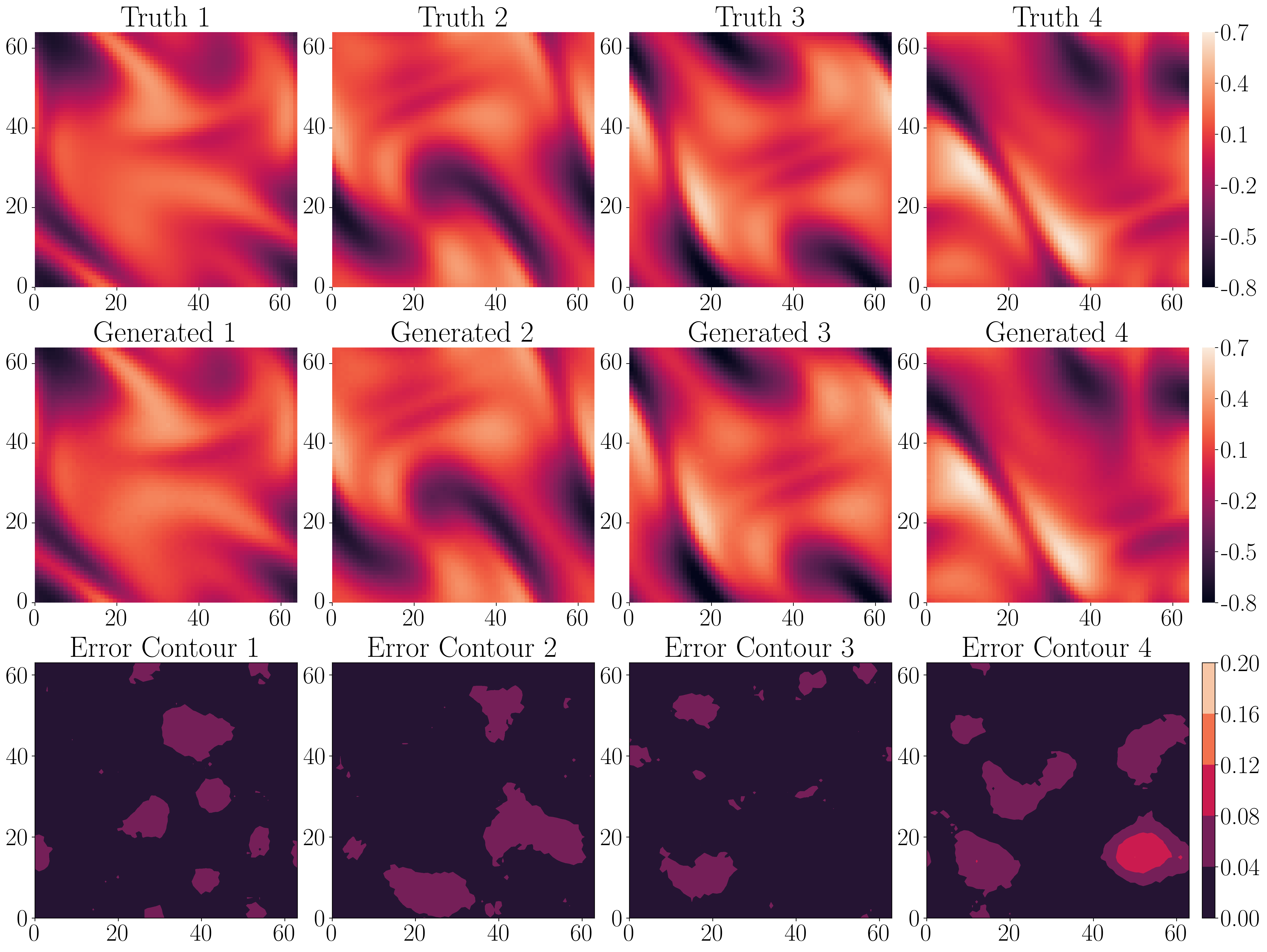}
    \caption{Physical-space conditional generation at $64\times64$ resolution. \textbf{Top row:} Ground-truth closure terms $H(\mathrm{x}, t)$. \textbf{Middle row:} Generated samples $\hat{H}(\mathrm{x}, t)$ from the P-CDM. \textbf{Bottom row:} Absolute error fields $|H - \hat{H}|$.}
    \label{fig: physical-space-CDM}
\end{figure}

Utilizing an adaptive sampling schedule (Eq.~\eqref{eqn: adaptive_scheme}) on an NVIDIA RTX 4090 GPU, the P-CDM requires approximately 2.64 seconds to generate 1000 closure term samples. This baseline achieves a mean squared error of $D_\text{MSE} = 9.77 \times 10^{-4}$ and relative Frobenius norm error of $D_\text{RE} = 1.31 \times 10^{-1}$. Figure~\ref{fig: physical-space-CDM} illustrates the qualitative performance of this approach, showing ground truth closure terms, corresponding generated samples, and their absolute error fields.

To improve sampling efficiency of physical-space diffusion models, we employ convolutional autoencoders to compress the high-dimensional discretizations of both the vorticity field $\omega$ and the term $H$ into lower-dimensional latent representations:
\begin{equation}
    \begin{aligned}
    z^H &= \mathcal{E}_H(H), & \hat{H} &= \mathcal{D}_H(z^H), \\
    z^\omega &= \mathcal{E}_\omega(\omega), & \hat{\omega} &= \mathcal{D}_\omega(z^\omega),
    \end{aligned}
\end{equation}
where the encoders $\mathcal{E}_H$ and $\mathcal{E}_\omega$ map $64 \times 64$ fields to $16 \times 16$ latent representations, achieving 16$\times$ dimensional reduction. The effectiveness of this compression is quantified in Table~\ref{tab: decoupled_AE_error_metrics}, which demonstrates that errors of the reconstructed fields via the decoder are small and thus the autoencoders preserve a major amount of information contained in the original fields.

\begin{table}[!htbp]
\centering
\caption{Autoencoder reconstruction performance using separate training for closure terms and resolved states. The low reconstruction errors confirm that the $16 \times 16$ latent space effectively captures the essential characteristics while eliminating redundant information.}
\begin{adjustbox}{width=0.3\linewidth}
\begin{tabular}{lcc}
\toprule
\textbf{Field} & \textbf{Metric} & \textbf{Value} \\
\midrule
\multirow{2}{*}{$H(\mathrm{x}, t)$} & $D_\text{MSE}$ & 1.158e-05 \\
 & $D_\text{RE}$ & 1.421e-02 \\
\midrule
\multirow{2}{*}{$\omega(\mathrm{x}, t)$} & $D_\text{MSE}$ & 4.306e-06 \\
 & $D_\text{RE}$ & 1.738e-03 \\
\bottomrule
\end{tabular}
\end{adjustbox}
\label{tab: decoupled_AE_error_metrics}
\end{table}

\begin{figure}[!htbp]
    \centering
    \includegraphics[width=\linewidth]{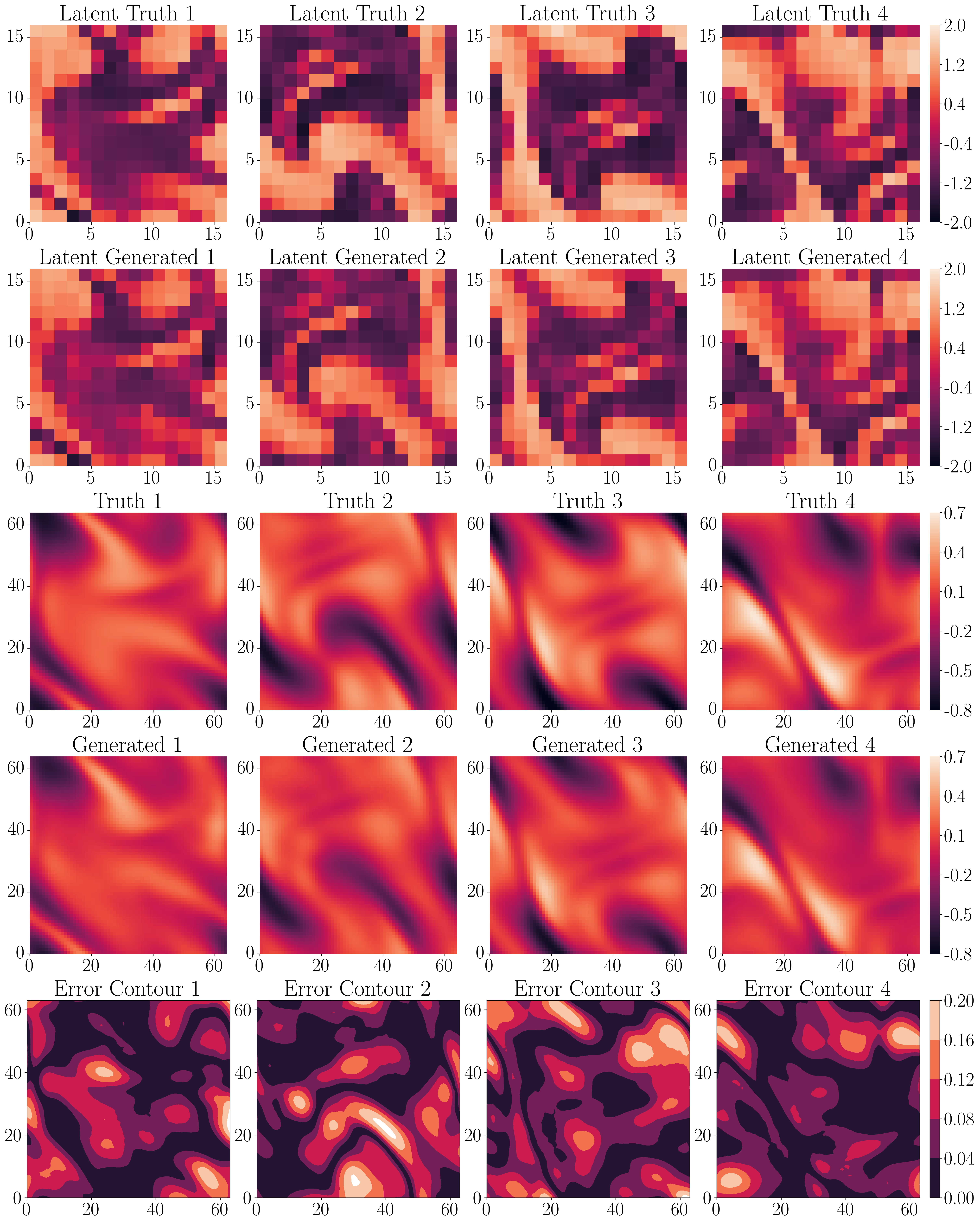}
    \caption{Generated samples from the latent space conditional diffusion model using conventional two-phase training. \textbf{First row:} Encoded ground truth latent representations $z^H$. \textbf{Second row:} Generated latent samples $\hat{z}^H$. \textbf{Third row:} Ground truth closure terms $H$. \textbf{Fourth row:} Decoded closure terms $\hat{H}$. \textbf{Fifth row:} Absolute error fields $|H - \hat{H}|$.}
    \label{fig: ModelWithoutJoint}
\end{figure}

In this example, the latent-space conditional diffusion model (L-CDM) is trained with the conventional two-phase training approach, where autoencoders and the diffusion model are optimized separately. It is worth noting that the L-CDM demonstrates remarkable computational efficiency, generating 1000 physical-space samples in approximately 0.48 seconds, which demonstrates a 5.5$\times$ acceleration compared to the P-CDM baseline. This efficiency stems from operating in the lower-dimensional latent space, which substantially reduces the computational complexity of the reverse diffusion process that generates samples based on a trained diffusion model. Figure~\ref{fig: ModelWithoutJoint} visualizes the generated samples, showing both the latent representations and decoded physical fields. Quantitatively, the L-CDM yields latent samples with error metrics $D_\text{MSE} = 1.23 \times 10^{-1}$ and $D_\text{RE} = 3.53 \times 10^{-1}$, while the decoded physical-space closure terms $H$ exhibit $D_\text{MSE} = 4.80 \times 10^{-3}$ and $D_\text{RE} = 2.91 \times 10^{-1}$.

It can be observed that, while the separately trained L-CDM achieves substantial acceleration in sample generation, the accuracy of those generated samples is noticeably worse than the physical-space baseline. This accuracy gap suggests a limitation of the conventional two-phase training approach, i.e., the optimal latent space solely dictated by the reconstruction accuracy of autoencoders can lead to a latent distribution that is challenging for existing diffusion model techniques. Instead of developing more sophisticated diffusion model techniques to handle those challenging latent distributions, this work aims to study an alternative approach that jointly trains the autoencoder and the latent diffusion model, with which the autoencoder not only guarantees small reconstruction errors but also ensures a latent space that supports satisfactory performance with existing diffusion model techniques. Detailed results of the jointly trained L-CDM are presented in the following section.

\subsection{Diffusion model in a jointly trained latent space}\label{ssec: joint_results}
In this section, we compare the conventional two-phase training approach with our proposed integrated training paradigm, where autoencoder and diffusion components are jointly optimized using the unified loss function defined in Eq.~\eqref{eqn: joint_loss_full}. With the joint training framework, autoencoders and the latent diffusion model are simultaneously optimized, allowing latent representations to evolve in response to both reconstruction and generative objectives. Table~\ref{tab: joint_AE_error_metrics} quantifies the reconstruction performance of jointly trained autoencoders, confirming effective latent representation learning despite the additional constraints imposed by concurrent diffusion model training.

\begin{table}[!htbp]
\centering
\caption{Reconstruction performance with jointly trained autoencoders. Errors increase slightly compared to the separately trained autoencoders (see Table~\ref{tab: decoupled_AE_error_metrics}).}
\begin{adjustbox}{width=0.3\linewidth}
\begin{tabular}{lcc}
\toprule
\textbf{Field} & \textbf{Metric} & \textbf{Value} \\
\midrule
\multirow{2}{*}{$H(\mathrm{x}, t)$} & $D_\text{MSE}$ & 7.965e-05 \\
 & $D_\text{RE}$ & 3.694e-02 \\
\midrule
\multirow{2}{*}{$\omega(\mathrm{x}, t)$} & $D_\text{MSE}$ & 6.429e-06 \\
 & $D_\text{RE}$ & 2.595e-03 \\
\bottomrule
\end{tabular}
\end{adjustbox}
\label{tab: joint_AE_error_metrics}
\end{table}

\begin{figure}[!htbp]
    \centering
    \includegraphics[width=\linewidth]{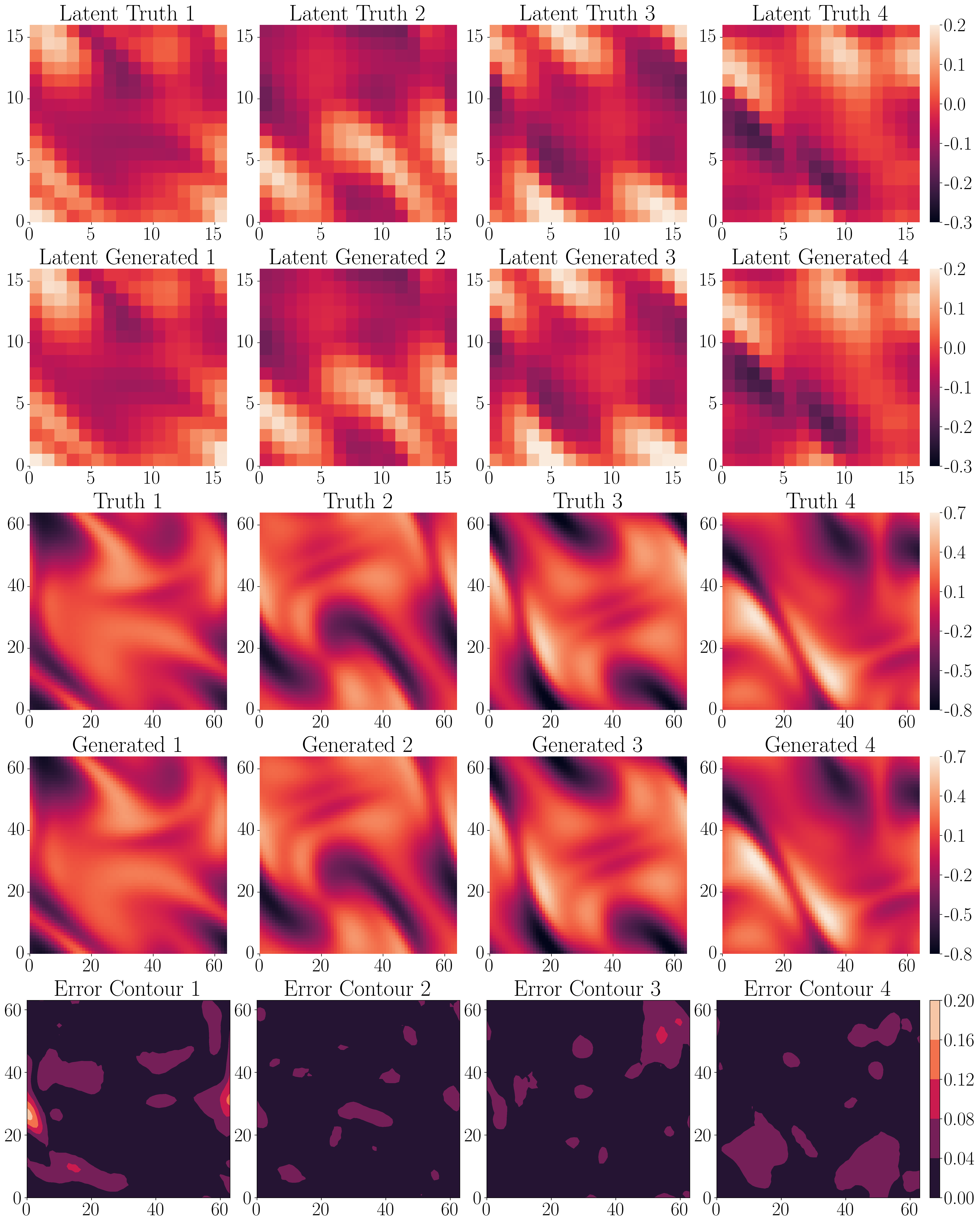}
    \caption{Generation results with jointly trained models. \textbf{First row:} Encoded ground truth $z^H$. \textbf{Second row:} Generated latent samples $\hat{z}^H$. \textbf{Third row:} Ground truth closure terms $H$. \textbf{Fourth row:} Decoded samples $\hat{H}$. \textbf{Fifth row:} Absolute error fields $|H - \hat{H}|$. Joint training produces more accurate latent representations and physical reconstructions compared to two-phase training (Figure~\ref{fig: ModelWithoutJoint}), with error magnitudes noticeably reduced across all samples.}
    \label{fig: ModelWithJoint}
\end{figure}

Figure~\ref{fig: ModelWithJoint} visualizes the generated samples using the jointly trained model. Specifically, the joint L-CDM produces latent samples with $D_\text{MSE} = 9.67 \times 10^{-5}$ and $D_\text{RE} = 1.24 \times 10^{-1}$, while the corresponding decoded physical fields exhibit $D_\text{MSE} = 7.40 \times 10^{-4}$ and $D_\text{RE} = 1.15 \times 10^{-1}$. This performance is comparable to that of the physical-space model while maintaining the computational efficiency of latent diffusion, with generation times of approximately 0.49 seconds per 1000 samples.

\begin{table}[!htbp]
\centering
\caption{Performance comparison across model architectures and training methodologies. Bold values indicate best performance in each category. Joint training yields a substantial improvement in both latent and physical space accuracy while maintaining the computational advantages of latent diffusion.}
\begin{adjustbox}{width=\linewidth}
\begin{tabular}{lccccccc}
\toprule
\multirow{2}{*}{\textbf{Model}} 
& \multicolumn{2}{c}{\textbf{Reconstruction}} 
& \multicolumn{2}{c}{\makecell{\textbf{Latent space}\\\textbf{generation}}} 
& \multicolumn{2}{c}{\makecell{\textbf{Physical space}\\\textbf{generation}}} 
& \textbf{Cost} \\
\cmidrule(lr){2-3} \cmidrule(lr){4-5} \cmidrule(lr){6-7}
 & $D_\text{MSE}$ & $D_\text{RE}$ & $D_\text{MSE}$ & $D_\text{RE}$ & $D_\text{MSE}$ & $D_\text{RE}$ & (s/1000) \\
\midrule
P-CDM & -- & -- & -- & -- & 9.766e-04 & 1.314e-01 & 2.64 \\
Two-phase L-CDM & \textbf{1.158e-05} & \textbf{1.421e-02} & 1.234e-01 & 3.527e-01 & 4.803e-03 & 2.905e-01 & 0.48 \\
Joint L-CDM & 7.965e-05 & 3.694e-02 & \textbf{9.674e-05} & \textbf{1.243e-01} & \textbf{7.398e-04} & \textbf{1.146e-01} & 0.49 \\
\bottomrule
\end{tabular}
\end{adjustbox}
\label{tab: conclude}
\end{table}

Table~\ref{tab: conclude} presents a systematic comparison of all three approaches: P-CDM, two-phase L-CDM, and joint L-CDM. While separately-trained autoencoders in the conventional L-CDM achieve superior reconstruction performance, this pure-reconstruction-oriented latent space proves suboptimal for conditional diffusion modeling. In contrast, the jointly trained autoencoder with the latent diffusion model produces latent representations that simultaneously support both reconstruction and generation performances, leading to physical-space generation accuracy that is on par with the baseline P-CDM while still exploiting the computational efficiency of latent approaches.

\begin{figure}[!htbp]
    \centering
    \includegraphics[width=\linewidth]{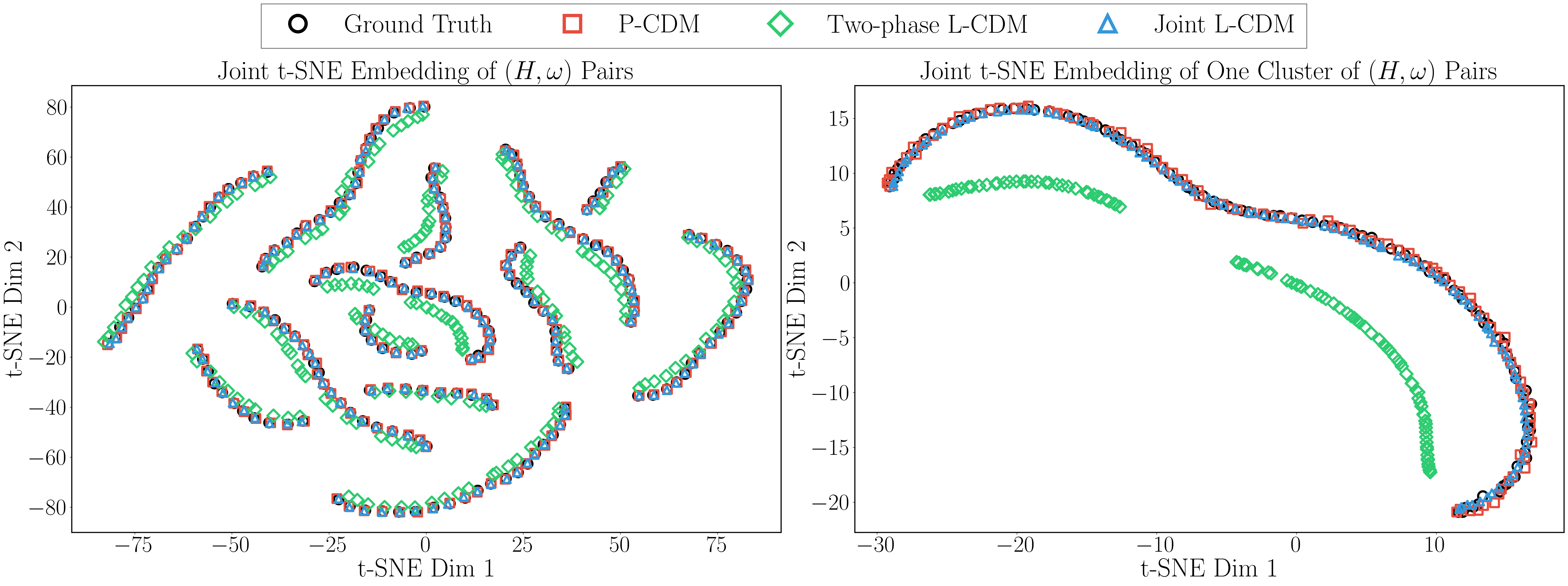}
    \caption{t-SNE visualization of $(H, \omega)$ pairs from ground truth and generated samples. \textbf{Left:} Ten distinct clusters corresponding to different test trajectories. \textbf{Right:} One representative cluster corresponding to one test trajectory. }
    \label{fig: tsne_full}
\end{figure}

To further visualize the overall performance differences of different trained models, we project the samples from the joint distribution $p(H,\omega)$ produced by each model into a two-dimensional space using t-SNE (Fig.~\ref{fig: tsne_full}), illustrating how the results of joint L-CDM better align generated samples with the ground truth distribution, compared to the results from the conventional two-phase L-CDM.

As shown in Fig.~\ref{fig: tsne_full}, samples from conventional L-CDM (green) exhibit substantial distributional divergence from ground truth samples (black), while both P-CDM (red) and joint L-CDM (blue) maintain good overall alignment. Quantitative analysis of a representative cluster (Fig.~\ref{fig: tsne_full}, right panel) confirms this observation, with two-phase L-CDM exhibiting substantially higher relative error ($D_\text{RE} = 3.23 \times 10^{-1}$) compared to joint L-CDM ($D_\text{RE} = 1.19 \times 10^{-1}$) and P-CDM ($D_\text{RE} = 1.22 \times 10^{-1}$). 

\begin{figure}[tbp]
    \centering
    \includegraphics[width=0.6\linewidth]{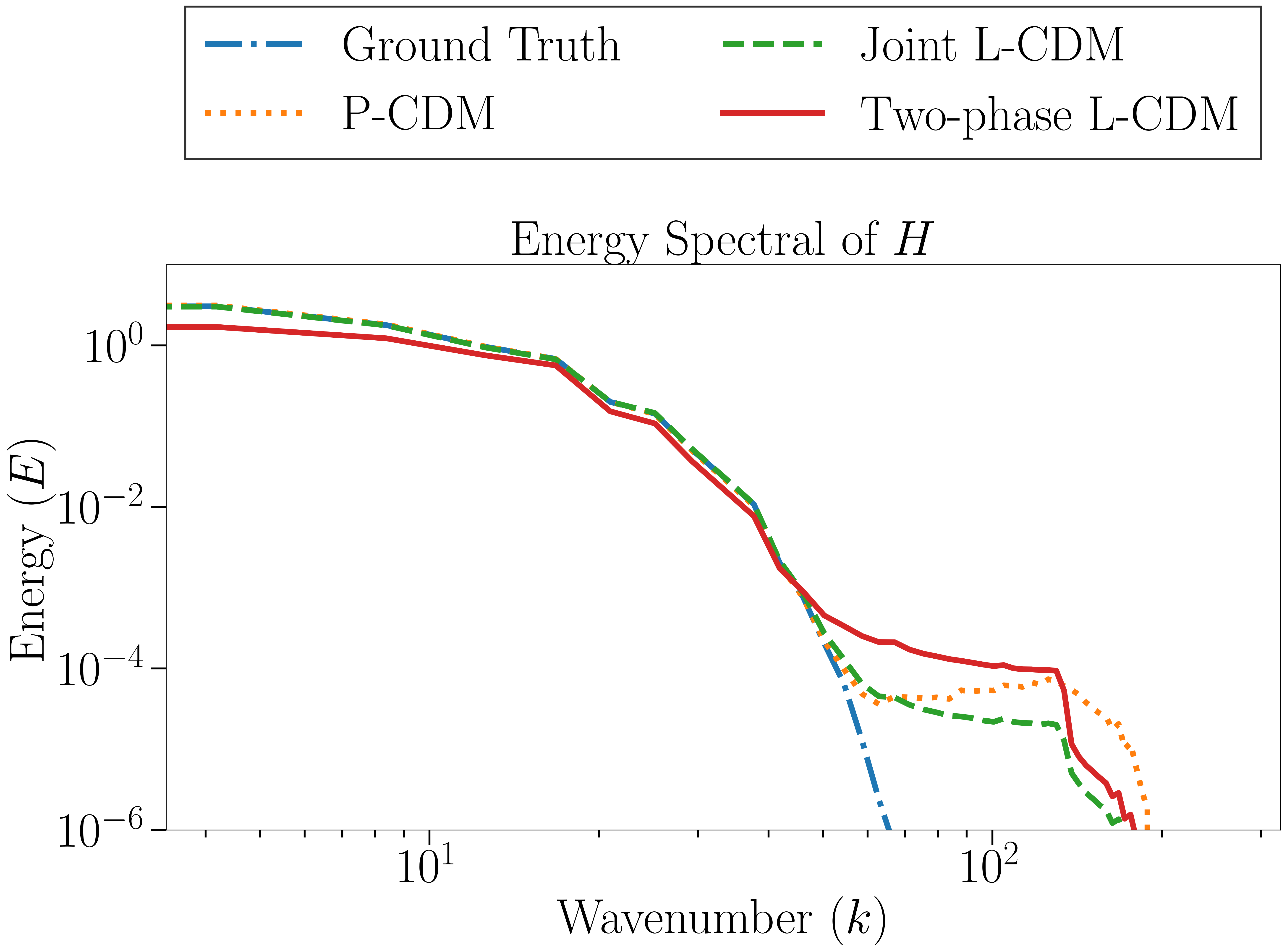}
     \caption{Mean energy spectra of the ground-truth closure term \(H\) and of closure fields generated by P-CDM, two-phase L-CDM, and joint L-CDM.}
    \label{fig: combined_energy_spectrum}
\end{figure}

Figure~\ref{fig: combined_energy_spectrum} provides additional comparison of different training methodologies using energy spectrum analysis in both the physical space and the latent space. The energy spectrum is defined as:
\begin{align}\label{eqn: energy_spectrum}
    E(k, t) = \frac{1}{2} \left |\hat{H}(k, t) \right |^2,
\end{align}
where $\hat{H}(k, t) = \mathcal{F}(H(\mathrm{x}, t))$ denotes the Fourier transform of the field $H(\mathrm{x}, t)$ with $k$ as the wavenumber. $| \cdot |$ evaluates the magnitude of a complex number. This spectral representation characterizes how kinetic energy is distributed across different spatial scales, with low wavenumbers corresponding to large-scale coherent structures and high wavenumbers representing fine-scale fluctuations. It is worth noting that the two-phase L-CDM exhibits a systematic spectral distortion: it underestimates energy at low wavenumbers while producing an elevated high-wavenumber tail. In contrast, P-CDM and Joint L-CDM match the ground-truth spectrum more closely over the low- and intermediate-wavenumber ranges, with the main discrepancies appearing in the high-wavenumber tail. We note that these discrepancies are related to spectral bias in neural-operator architectures~\cite{khodakarami2025mitigating}. Some recent works~\cite{wang2024multi,khodakarami2025mitigating} have explored methods to reduce spectral biases for general neural networks and specifically for neural operators, and leveraging those methods to improve the results within the high-wavenumber tail remains an important direction among future extensions of this work.

The stochastic nature of diffusion sampling introduces intrinsic variability in generated closure terms. To characterize this behavior, we also analyze the performance of multiple samples ($N_e=1000$) generated for fixed conditioning variables and confirm that the joint L-CDM outperforms the two-phase L-CDM in terms of the ensemble mean values. Table~\ref{tab: ensemble_metrics} presents the range of performance metrics observed across an ensemble of samples, along with the accuracy of ensemble means.

Table~\ref{tab: ensemble_metrics} also allows for an evaluation of how well the models represent the prescribed stochasticity, quantified by the spatially-averaged standard deviation (Field Std.). For our problem setup, this quantity has a known analytical reference value of $\approx 0.02$, as derived in~\ref{sec: ablation_stochastic}. Both the P-CDM and our Joint L-CDM reproduce this target with high fidelity. In contrast, the Two-phase L-CDM, consistent with its poor mean-field performance, fails to capture this physical variance. This metric is critical, as accurately capturing the uncertainty level is essential for preserving the system’s physical statistics in \textit{a posteriori} ensemble simulations.

These combined results establish that the proposed joint training strategy is successful, yielding a model that accurately and efficiently captures both the mean behavior and the stochastic nature of the closure.~\ref{sec: ablation_stochastic} provides a further ablation study that demonstrates the fundamental necessity of a stochastic, rather than deterministic, model for this task.

These results also highlight a critical nuance in latent variable modeling regarding data efficiency. While compressing high-dimensional fields into a lower-dimensional latent space theoretically mitigates the curse of dimensionality and reduces data requirements, dimensionality reduction alone does not guarantee a simpler learning problem. As seen with the Two-phase L-CDM, a latent space optimized solely for reconstruction can yield a highly irregular and topologically complex distribution. This complexity negates the benefits of the reduced dimension, making the conditional distribution $p(z^U|z^V)$ difficult to learn with finite data. The superior performance of the Joint L-CDM confirms that the $\mathcal{L}_\text{joint}$ objective is essential for ensuring that the reduction in dimensionality translates into a genuine gain in data efficiency and generative robustness.

\begin{table}[!htbp]
\centering
\caption{Ensemble sampling performance over 1000 samples generated with the same conditioning input. Error metrics are grouped into mean of per-sample errors and error of the ensemble-averaged prediction. Field Std. is the spatially-averaged standard deviation of the generated ensemble, indicating the magnitude of modeled uncertainty. Values in gray denote $\pm$ two standard deviation over the test set instances. Bold values indicate best performance in each category.}
\begin{adjustbox}{width=\linewidth}
\begin{tabular}{lcccccr}
\toprule
\multirow{2}{*}{\textbf{Model}} 
& \multicolumn{2}{c}{\textbf{Mean of per-sample errors}} 
& \multicolumn{2}{c}{\textbf{Error of ensemble mean}} 
& \multirow{2}{*}{\textbf{Field Std.}} \\
\cmidrule(lr){2-3} \cmidrule(lr){4-5}
& $D_\text{RE}^{\text{sample}}$ & $D_\text{MSE}^{\text{sample}}$ 
& $D_\text{RE}^{\text{ens}}$ & $D_\text{MSE}^{\text{ens}}$ 
& \\
\midrule
P-CDM 
& 1.148e-01 \textcolor{gray}{\scriptsize$\pm$ 2.541e-02} 
& 8.451e-04 \textcolor{gray}{\scriptsize$\pm$ 3.729e-04}
& \textbf{6.851e-02} & \textbf{2.974e-04} 
& 1.8890e-02 \textcolor{gray}{\scriptsize$\pm$ 2.711e-03} \\
Two-phase L-CDM 
& 3.401e-01 \textcolor{gray}{\scriptsize$\pm$ 3.642e-02} 
& 7.350e-03 \textcolor{gray}{\scriptsize$\pm$ 1.579e-03} 
& 3.225e-01 & 6.585e-03 
& 4.1005e-02 \textcolor{gray}{\scriptsize$\pm$ 6.873e-03} \\
Joint L-CDM 
& \textbf{9.702e-02} \textcolor{gray}{\scriptsize$\pm$ 1.703e-02} 
& \textbf{6.012e-04} \textcolor{gray}{\scriptsize$\pm$ 2.122e-04}
& 7.663e-02 & 3.722e-04
& \textbf{2.1202e-02} \textcolor{gray}{\scriptsize$\pm$ 1.658e-03} \\
\bottomrule
\end{tabular}
\end{adjustbox}
\label{tab: ensemble_metrics}
\end{table}

\subsection{Numerical simulations of the vorticity using the trained models}\label{sec: surrogate}

To evaluate the practical efficacy of our latent diffusion closure modeling framework, we integrate the trained models into a numerical solver for the 2-D Navier--Stokes system in Eq.~\eqref{2dNS}. More specifically, we conduct numerical simulations of the vorticity field $\omega(\mathrm{x}, t)$ over $t \in [30, 50]$ with temporal resolution $\Delta t = 10^{-3}$, using the same pseudo-spectral and Crank-Nicolson methods described in \ref{sec: data_generation}. The following conditions are prescribed:

\begin{itemize}
    \item Initial vorticity field: $\omega(\mathrm{x}, t_0=30)$ from the high-fidelity dataset
    \item Deterministic forcing: $f(\mathrm{x}) = 0.1 (\sin(2\pi(x+y)) + \cos(2\pi(x+y)))$
    \item Viscosity coefficient: $\nu = 10^{-3}$
\end{itemize}

To optimize computational efficiency while maintaining physical accuracy, closure terms are generated every 5 physical timesteps rather than at every integration step. The reverse diffusion process uses an adaptive schedule with 10 timesteps and maximum sampling time $T_\text{sample}=0.1$, as described in Section~\ref{sssec: CDM}. For implementation, we evaluate two stochastic closure models (P-CDM and joint L-CDM) using two integration strategies:

    

\begin{enumerate}
    \item \textbf{Per-sample simulation}: At each evaluation time step, a single sample $\hat{H}(\mathbf{x}, t)$ is drawn from the conditional distribution $p(H\mid\omega)$. We independently simulate $N_e=1000$ such trajectories, and report the mean and standard deviation of the resulting prediction errors.

    \item \textbf{Ensemble-mean simulation}: At each evaluation time step, multiple samples $\{\hat{H}_i(\mathbf{x}, t)\}_{i=1}^{N_e}$ are drawn from $p(H\mid\omega)$, and their ensemble average $\bar{H}(\mathbf{x}, t) = \frac{1}{N_e} \sum_{i=1}^{N_e} \hat{H}_i(\mathbf{x}, t)$ is used. Here, $N_e = 1000$.
\end{enumerate}

\begin{figure}[!htb]
    \centering
    \includegraphics[width = \linewidth]{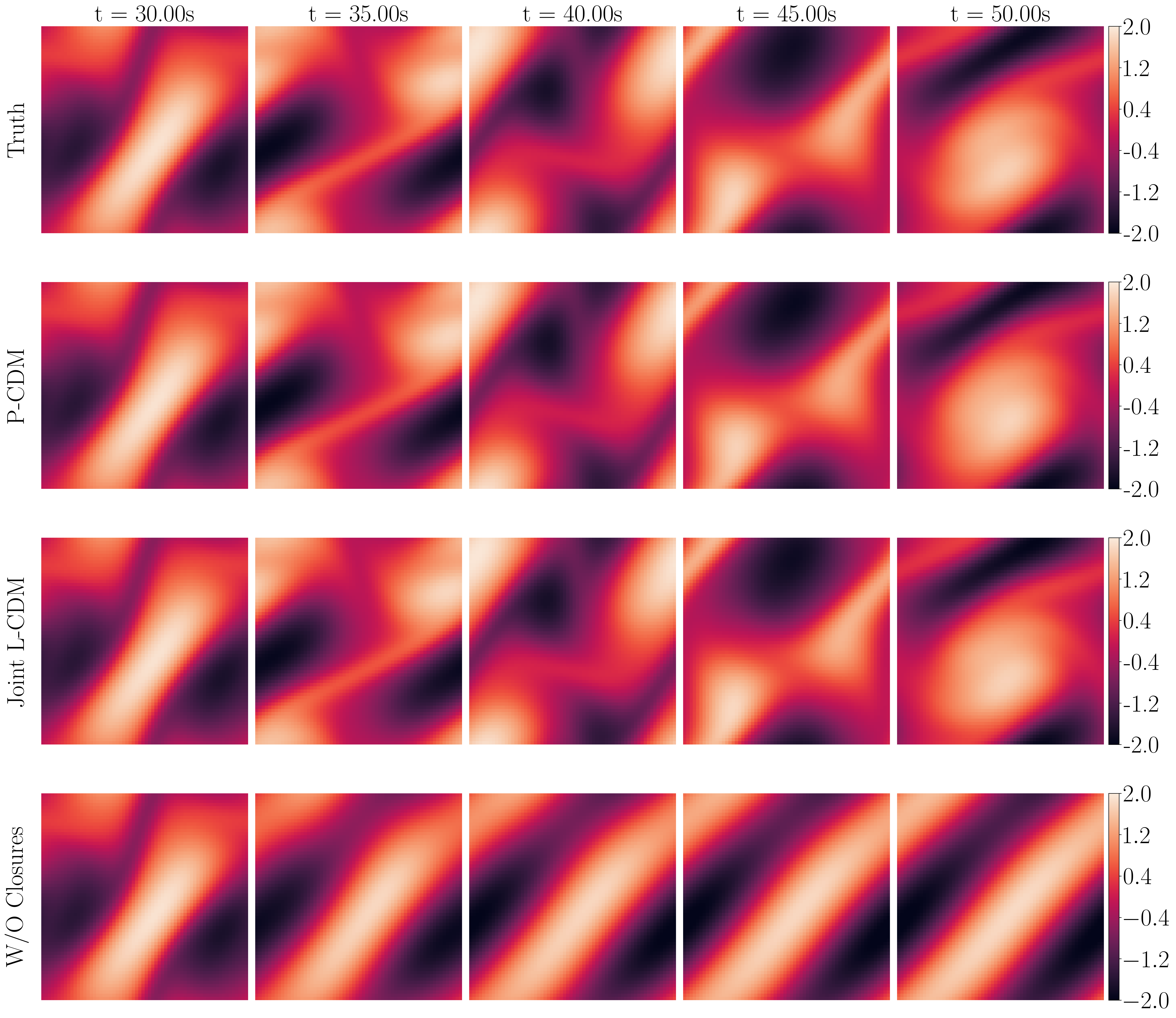}
    \caption{Temporal evolution of vorticity fields over a 20-second simulation period. \textbf{First row:} Ground truth vorticity $\omega$ from high-fidelity simulation. \textbf{Second row:} Simulated vorticity using P-CDM ensemble-mean closure. \textbf{Third row:} Simulated vorticity using joint L-CDM ensemble-mean closure. \textbf{Fourth row:} Simulated vorticity with closure terms neglected.}
    \label{fig: surrogate_H}
\end{figure}

\begin{figure}[!htbp]
    \centering
    \includegraphics[width=\linewidth]{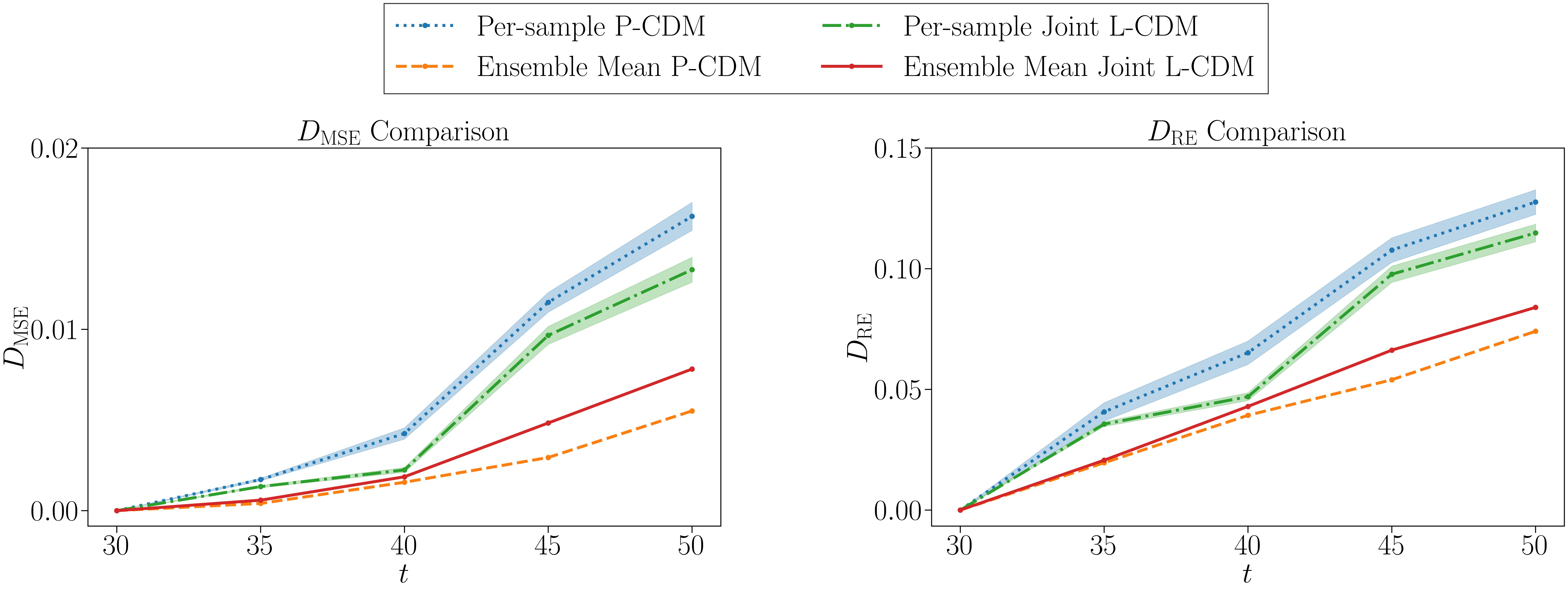}
    \caption{Temporal evolution of simulation errors for different closure modeling strategies. \textbf{Left:} Mean squared error ($D_\text{MSE}$). \textbf{Right:} Relative Frobenius norm error ($D_\text{RE}$). All closure models significantly outperform the uncorrected simulation, with ensemble strategies showing consistent advantages over single-trajectory approaches. The ensemble P-CDM achieves marginally lower errors than ensemble joint L-CDM, but at a substantially higher computational cost.}
    \label{fig: MSE_RE_Comparison}
\end{figure}

Figure~\ref{fig: surrogate_H} presents a comparative visualization of vorticity fields at different simulation times. Three configurations are shown: (i) ground truth high-fidelity solution, (ii) simulation that neglects the closure term, and (iii) simulation with the joint L-CDM ensemble-mean closure. The uncorrected simulation exhibits progressive degradation in solution fidelity, with substantial deviations in both the overall structure and the intensity becoming apparent beyond $t=35$. In contrast, the joint L-CDM closure simulation maintains a good agreement with the reference solution throughout the whole time integration period. Table~\ref{tab: simulation_metrics} and Fig.~\ref{fig: MSE_RE_Comparison} quantitatively confirm the good simulation performance of using the diffusion models to generate the stochastic closure term. The uncorrected model rapidly deviates from the ground truth, reaching a relative error of 81.6\% by $t=50$, indicating that the closure term is important in terms of achieving a good quantitative agreement of the simulated results with the true system. In contrast, both closure models maintain significantly better accuracy throughout the simulation period, with relative errors remaining below 15\% for single-trajectory simulations and below 10\% for ensemble simulations. 

\begin{table}[!tb]
\centering
\caption{Simulation performance over a 20-second integration. Reported are per-sample mean errors and 2-standard-deviation bands over 1000 independently simulated trajectories, as well as ensemble mean prediction errors based on 1000-sample averages. The reported time cost is defined as the average wall-clock time per trajectory for per-sample simulations, and the total time required to generate and average 1000 samples for ensemble simulations.}
\begin{adjustbox}{width=\linewidth}
\begin{tabular}{lccccccccc}
\toprule
\multirow{2}{*}{\textbf{Model}} 
& \multirow{2}{*}{\textbf{Strategy}} 
& \multirow{2}{*}{\textbf{Cost (s)}} 
& \multirow{2}{*}{\textbf{Metric}} 
& \multirow{2}{*}{\textbf{Closure term}} 
& \multicolumn{5}{c}{\textbf{Vorticity field error at time}} \\
\cmidrule(lr){6-10}
& & & & & \textbf{t=30} & \textbf{t=35} & \textbf{t=40} & \textbf{t=45} & \textbf{t=50} \\
\midrule

\multirow{2}{*}{W/O closure} 
& \multirow{2}{*}{Per-sample} & \multirow{2}{*}{2.12} 
& $D_\text{RE}$ & -- 
& 0 & 5.35e-01 & 6.25e-01 & 7.08e-01 & 8.16e-01 \\
& & & $D_\text{MSE}$ & -- 
& 0 & 4.16e-01 & 4.53e-01 & 6.93e-01 & 8.58e-01 \\
\midrule

\multirow{6}{*}{P-CDM} 
& \multirow{4}{*}{Per-sample} & \multirow{4}{*}{176.54} 
& $D_\text{RE}$ & 1.20e-01 
& 0 & 4.07e-02 & 6.52e-02 & 1.08e-01 & 1.28e-01 \\
& & & \textcolor{gray}{\scriptsize ±2std} & \textcolor{gray}{\scriptsize$\pm$ 1.02e-02} 
& \textcolor{gray}{\scriptsize$\pm$ 0} & \textcolor{gray}{\scriptsize$\pm$ 7.59e-03} & \textcolor{gray}{\scriptsize$\pm$ 9.74e-03} & \textcolor{gray}{\scriptsize$\pm$ 1.02e-02} & \textcolor{gray}{\scriptsize$\pm$ 1.02e-02} \\
& & & $D_\text{MSE}$ & 8.00e-04 
& 0 & 1.72e-03 & 4.26e-03 & 1.15e-02 & 1.62e-02 \\
& & & \textcolor{gray}{\scriptsize ±2std} & \textcolor{gray}{\scriptsize$\pm$ 1.43e-04} 
& \textcolor{gray}{\scriptsize$\pm$ 0} & \textcolor{gray}{\scriptsize$\pm$ 1.43e-04} & \textcolor{gray}{\scriptsize$\pm$ 6.26e-04} & \textcolor{gray}{\scriptsize$\pm$ 1.09e-03} & \textcolor{gray}{\scriptsize$\pm$ 1.55e-03} \\
\cmidrule(lr){2-10}

& \multirow{2}{*}{Ensemble} & \multirow{2}{*}{8652.49} 
& $D_\text{RE}$ & 6.35e-02 
& 0 & 1.96e-02 & 3.93e-02 & 5.40e-02 & 7.41e-02 \\
& & & $D_\text{MSE}$ & 2.47e-04 
& 0 & 3.99e-04 & 1.57e-03 & 2.93e-03 & 5.50e-03 \\
\midrule

\multirow{6}{*}{Joint L-CDM} 
& \multirow{4}{*}{Per-sample} & \multirow{4}{*}{140.67} 
& $D_\text{RE}$ & 1.05e-01 
& 0 & 3.56e-02 & 4.70e-02 & 9.77e-02 & 1.15e-01 \\
& & & \textcolor{gray}{\scriptsize ±2std} & \textcolor{gray}{\scriptsize$\pm$ 7.33e-03} 
& \textcolor{gray}{\scriptsize$\pm$ 0} & \textcolor{gray}{\scriptsize$\pm$ 2.03e-03} & \textcolor{gray}{\scriptsize$\pm$ 3.26e-03} & \textcolor{gray}{\scriptsize$\pm$ 6.76e-03} & \textcolor{gray}{\scriptsize$\pm$ 7.33e-03} \\
& & & $D_\text{MSE}$ & 2.18e-04 
& 0 & 1.33e-03 & 2.25e-03 & 9.67e-03 & 1.33e-02 \\
& & & \textcolor{gray}{\scriptsize ±2std} & \textcolor{gray}{\scriptsize$\pm$ 1.20e-05} 
& \textcolor{gray}{\scriptsize$\pm$ 0} & \textcolor{gray}{\scriptsize$\pm$ 1.20e-05} & \textcolor{gray}{\scriptsize$\pm$ 2.63e-04} & \textcolor{gray}{\scriptsize$\pm$ 9.94e-04} & \textcolor{gray}{\scriptsize$\pm$ 1.37e-03} \\
\cmidrule(lr){2-10}

& \multirow{2}{*}{Ensemble} & \multirow{2}{*}{1236.87} 
& $D_\text{RE}$ & 7.02e-02 
& 0 & 2.06e-02 & 4.29e-02 & 6.63e-02 & 8.40e-02 \\
& & & $D_\text{MSE}$ & 3.01e-04 
& 0 & 5.78e-04 & 1.87e-03 & 4.84e-03 & 7.81e-03 \\
\bottomrule
\end{tabular}
\end{adjustbox}
\label{tab: simulation_metrics}
\end{table}

In addition to the accuracy study, it is worth noting that the joint L-CDM achieves a $\sim 10\times$ acceleration in ensemble simulation compared to P-CDM (1,236s vs. 8,652s), with final relative error metrics differing by only 1 percentage point at the final timestep ($8.4\%$ vs. $7.4\%$). This dramatic speedup stems from performing the computationally intensive sampling process in a lower-dimensional latent space, rather than the full physical domain. The efficiency-accuracy balance achieved by the joint L-CDM makes it particularly well-suited for practical uncertainty quantification studies where large ensemble simulations are required. Its ability to generate physically consistent closure terms with an affordable computational overhead represents a significant advancement in stochastic closure modeling for complex dynamical systems.

\begin{figure}[!htbp]
    \centering
    \includegraphics[width=\linewidth]{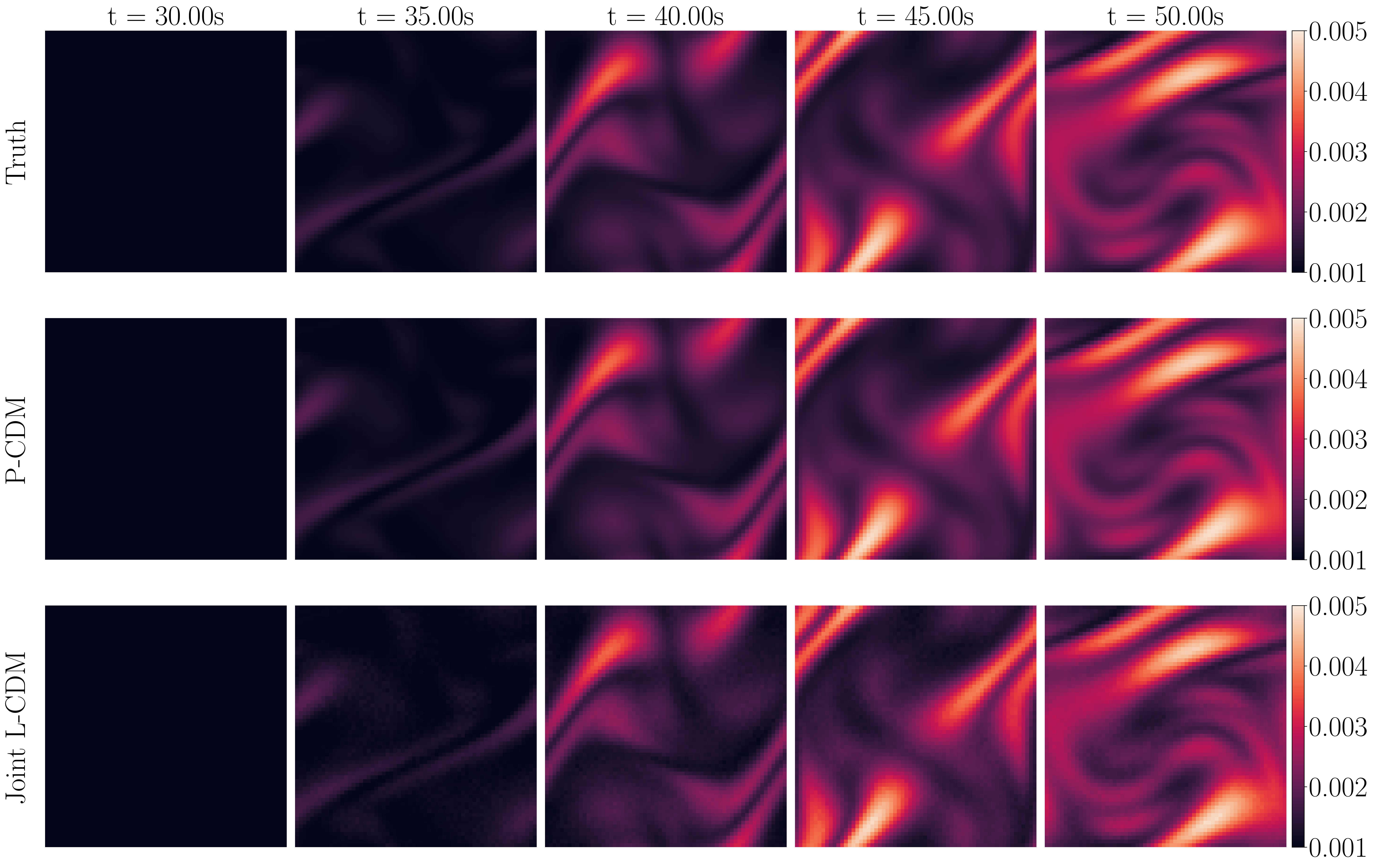}
    \caption{
        Spatial distribution of simulation uncertainty. The pixel-wise standard deviation is computed across an ensemble of 1000 stochastic simulations. The top row shows the ground truth variability, while the bottom two rows show the variability captured by the P-CDM and Joint L-CDM closure models respectively. The close agreement in both structure and magnitude demonstrates the model's ability to reproduce the physical uncertainty of the system.
    }
    \label{fig: closure_std}
\end{figure}

A key strength of the proposed stochastic closure, beyond predicting the mean resolved dynamics, is its ability to capture the intrinsic variability of the unresolved term. Figure~\ref{fig: closure_std} illustrates this by comparing the spatial standard deviation maps obtained from the learned stochastic ensembles against the ground truth. Both P-CDM and Joint L-CDM not only recover the mean state accurately, but also reproduce the spatial structure and magnitude of the uncertainty, indicating that the learned conditional distribution $p(H\mid\omega)$ represents physically meaningful variability rather than unstructured noise. In \ref{sec:LES_closure}, we further present additional results of applying the proposed stochastic closure modeling framework for learning the sub-grid scale models in under-resolved large eddy simulations.

\begin{figure}[!htbp]
    \centering
    \includegraphics[width=\linewidth]{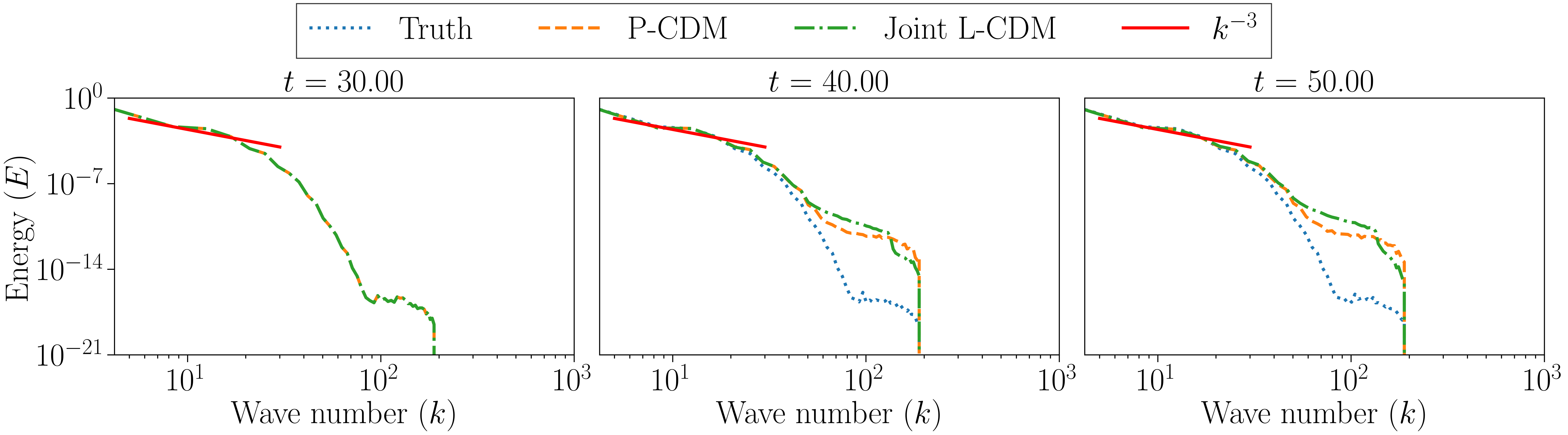}
    \caption{Comparison of energy spectra of 2-D Navier--Stokes equation among the ground truth, reduced-order model with closure $H$ generated with P-CDM, and reduced-order model with closure $H$ generated with Joint L-CDM at various times. Ensemble mean simulations are used. Only snapshots at times 30, 40, and 50 are shown since the energy spectra exhibit self-similarity during this period, making additional time points redundant.}
    \label{fig: TKE_3050}
\end{figure}

Figure~\ref{fig: TKE_3050} further compares the energy spectrum of the resolved vorticity among the ground truth simulation and the reduced-order model (ROM) with closure $H$ generated with P-CDM and Joint L-CDM at various times between 30s and 50s. It is evident that the simulations incorporating the trained closure models closely match the ground truth up to wavenumbers of order \(10^2\), demonstrating that most flow features are accurately captured with the added closure terms. As expected given the \(64\times64\) training resolution—and as shown in the spectral analysis in the left panel of Fig.~\ref{fig: combined_energy_spectrum}—the energy spectra deviate at higher wavenumbers. Nonetheless, the overall spectral shape follows the \(k^{-3}\) reference slope, indicating that the trained closures effectively preserve the forward enstrophy cascade characteristic of two-dimensional turbulence.

\section{Conclusion}
Modeling of complex multiscale dynamical systems in computational mechanics presents a fundamental challenge: the prohibitive computational cost of resolving all relevant scales. For systems without a clear scale separation, classical deterministic and local closure modeling methods can be too restrictive. This work addresses this challenge through a novel latent diffusion-based framework for stochastic and non-local closure modeling that balances accuracy and computational efficiency. In the conventional latent diffusion models, the latent space is determined purely based on the reconstruction performance, which may potentially lead to a latent distribution that is difficult to capture by standard diffusion model techniques. Compared to the conventional latent diffusion models, one of the key innovations in our approach is the joint training of autoencoders and diffusion models in the latent space, which facilitates an automatic adaptive tuning of the latent space to ensure strong diffusion-model performance. Numerical examples demonstrate that our jointly trained latent diffusion framework achieves $\sim 10\times$ computational acceleration in ensemble simulations while maintaining predictive accuracy comparable to physical-space diffusion models. By bridging the gap between representational expressiveness and computational feasibility, our approach enables practical ensemble-based uncertainty quantification that would be expensive with conventional techniques. 

While our numerical results demonstrate satisfactory performance on the test dataset, the framework's sensitivity to out-of-distribution data, such as extreme events not captured in the training set, remains an important area for future investigation. Future research directions also include incorporating physics-informed constraints directly into the latent representation and developing resolution-invariant autoencoder architectures that preserve the theoretical resolution-invariance properties of neural operators used to construct diffusion models, potentially enabling a unified framework capable of operating across arbitrary discretizations without retraining the model while maintaining the computational benefits of a properly discovered latent space. Finally, while this work focused on the study of a 2D numerical example, the framework is architecturally extensible to 3D systems. The primary challenge in such an extension would be the computational and memory costs associated with high-fidelity 3D data, which further underscores the necessity of the latent-space approach developed in this work. A comprehensive study of the proposed framework for large-scale 3D problems is an important direction for future research.

\section*{CRediT authorship contribution statement}
Xinghao Dong: Writing -- review \& editing, Writing -- original draft, Visualization, Validation, Software, Methodology, Investigation, Conceptualization; Huchen Yang: Writing -- review \& editing, Validation; Jin-Long Wu: Writing -- review \& editing, Supervision, Methodology, Funding acquisition, Conceptualization.

\section*{Data availability}
The data and trained models that support the findings of this study are available from the corresponding author upon reasonable request. The codes and examples that support the findings of this study are available at \url{https://github.com/AIMS-Madison/Latent_Diffusion_Closures}.

\section*{Declaration of competing interest}
The authors declare that they have no known competing financial interests or personal relationships that could have appeared to influence the work reported in this paper.

\section*{Acknowledgments}
X.D., H.Y. and J.W. are supported by the University of Wisconsin-Madison, Office of the Vice Chancellor for Research and Graduate Education with funding from the Wisconsin Alumni Research Foundation. X.D. and J.W. are also funded by the Office of Naval Research N00014-24-1-2391.

\clearpage

\appendix

\section{Equivalence of conditional score matching objectives}\label{sec:score_matching_proof}
This appendix provides a formal justification for the conditional score-matching objective function used in Eq.~\eqref{eqn:latent_conditionscorematching}. We first show the standard equivalence between Explicit Score Matching (ESM) and Denoising Score Matching (DSM) for the unconditional case, and then extend this to the conditional case to validate our objective.

\subsection{Equivalency between ESM and DSM (unconditional)}\label{ssec: proof_ESMDSM}
We begin with the unconditional case, using the latent variable $z^U$ for notation. The Explicit Score Matching (ESM) objective is:
\begin{equation}\label{eqn: ESM}
    \begin{aligned}
        J_\text{ESM}(\theta) &= \mathbb{E}_{z^U_\tau \sim p(z^U_\tau)} \left \| \nabla_{z^U_\tau} \log p(z^U_\tau) - s_\theta(\tau, z^U_\tau)\right \|^2_2 \\
        &= \mathbb{E}_{z^U_\tau \sim p(z^U_\tau)} \|s_\theta(\tau, z^U_\tau) \|^2_2 - 2H(\theta) + C_1
    \end{aligned}
\end{equation}
where $C_1 = \mathbb{E}_{z^U_\tau \sim p(z^U_\tau)} \| \nabla_{z^U_\tau} \log p(z^U_\tau) \|^2_2$ is a constant independent of $\theta$. Using integration by parts (or Green's first identity) and assuming mild boundary conditions, $H(\theta)$ can be rewritten:
\begin{equation}
    \begin{aligned}
        H(\theta) &= \mathbb{E}_{z^U_\tau \sim p(z^U_\tau)} \left [\left < \nabla_{z^U_\tau} \log p(z^U_\tau), s_\theta(\tau, z^U_\tau) \right > \right]\\
        &= \int p(z^U_\tau) \left < \frac{\nabla_{z^U_\tau}p(z^U_\tau)}{p(z^U_\tau)}, s_\theta \right > \mathrm{d}z^U_\tau \\
        &= \int \left < \nabla_{z^U_\tau} p(z^U_0) p(z^U_\tau \mid z^U_0) \mathrm{d}z^U_0, s_\theta \right > \mathrm{d}z^U_\tau \\
        &= \iint p(z^U_0) p(z^U_\tau \mid z^U_0) \left < \nabla_{z^U_\tau} \log p(z^U_\tau \mid z^U_0), s_\theta \right > \mathrm{d}z^U_0 \mathrm{d}z^U_\tau \\
        &= \mathbb{E}_{z^U_0 \sim p(z^U_0)} \mathbb{E}_{z^U_\tau \sim p(z^U_\tau \mid z^U_0)} \left[ \left < \nabla_{z^U_\tau}\log p(z^U_\tau \mid z^U_0), s_\theta(\tau, z^U_\tau) \right > \right].
    \end{aligned}
\end{equation}
The Denoising Score Matching (DSM) objective (from Eq.~\eqref{eqn:latent_ESMDSM}) is:
\begin{equation}\label{eqn: DSM}
    \begin{aligned}
        J_\text{DSM}(\theta) =& \mathbb{E}_{z^U_0 \sim p(z^U_0)} \mathbb{E}_{z^U_\tau \sim p(z^U_\tau \mid z^U_0)} \left \| \nabla_{z^U_\tau} \log p(z^U_\tau \mid z^U_0) - s_\theta(\tau, z^U_\tau)\right \|^2_2 \\
        =& \mathbb{E}_{z^U_\tau \sim p(z^U_\tau)} \|s_\theta(\tau, z^U_\tau) \|^2_2 \\
        &- 2 \mathbb{E}_{z^U_0 \sim p(z^U_0)} \mathbb{E}_{z^U_\tau \sim p(z^U_\tau \mid z^U_0)} \left[ \left < \nabla_{z^U_\tau}\log p(z^U_\tau \mid z^U_0), s_\theta \right > \right] + C_2,
    \end{aligned}
\end{equation}
where $C_2 = \mathbb{E}_{z^U_0 \sim p(z^U_0)} \mathbb{E}_{z^U_\tau \sim p(z^U_\tau \mid z^U_0)} \| \nabla_{z^U_\tau} \log p(z^U_\tau \mid z^U_0) \|^2_2$ is a constant.
Comparing the expanded forms, $J_\text{ESM}(\theta) = J_\text{DSM}(\theta) + C_1 - C_2$. Thus, minimizing $J_\text{ESM}$ is equivalent to minimizing $J_\text{DSM}$. This proves the DSM objective in Eq.~\eqref{eqn:latent_ESMDSM} is valid.

\subsection{Conditional score matching (CSM)}\label{ssec: CSM}
We now extend this to the conditional case to validate Eq.~\eqref{eqn:latent_conditionscorematching}. Our goal is to model the conditional score $\nabla_{z^U_\tau} \log p(z^U_\tau \mid z^V)$.
The forward diffusion process is independent of the condition $z^V$, meaning:
\begin{align}
    p(z^U_\tau \mid z^U_0, z^V) = p(z^U_\tau \mid z^U_0)
\end{align}
This is a key property. The "ideal" DSM objective for the conditional score $s_\theta(\tau, z^U_\tau, z^V)$ would be to minimize the expected error over the true conditional data distribution $p(z^U_0 \mid z^V)$:
\begin{equation}
    \begin{aligned}
        J_\text{CSM}(\theta) &= \mathbb{E}_{z^V \sim p(z^V)} \mathbb{E}_{z^U_0 \sim p(z^U_0 \mid z^V)} \mathbb{E}_{z^U_\tau \sim p(z^U_\tau \mid z^U_0, z^V)} \left[ \left \| \nabla_{z^U_\tau} \log p(z^U_\tau \mid z^U_0) - s_\theta(\tau, z^U_\tau, z^V)\right \|^2_2 \right] \\
        &= \mathbb{E}_{z^V \sim p(z^V)} \mathbb{E}_{z^U_0 \sim p(z^U_0 \mid z^V)} \mathbb{E}_{z^U_\tau \sim p(z^U_\tau \mid z^U_0)} \left[ \left \| \nabla_{z^U_\tau} \log p(z^U_\tau \mid z^U_0) - s_\theta(\tau, z^U_\tau, z^V)\right \|^2_2 \right]
    \end{aligned}
\end{equation}
By the definition of expectation, we can combine the outer two expectations:
\begin{align}
    \mathbb{E}_{z^V \sim p(z^V)} \left[ \mathbb{E}_{z^U_0 \sim p(z^U_0 \mid z^V)} [\dots] \right] = \mathbb{E}_{(z^U_0, z^V) \sim p(z^U_0, z^V)} [\dots]
\end{align}
Substituting this back, we get:
\begin{equation}
    \begin{aligned}
        J_\text{CSM}(\theta) = \mathbb{E}_{(z^U_0, z^V) \sim p(z^U_0, z^V)} \mathbb{E}_{z^U_\tau \sim p(z^U_\tau \mid z^U_0)} \left[ \left \| \nabla_{z^U_\tau} \log p(z^U_\tau \mid z^U_0) - s_\theta(\tau, z^U_\tau, z^V)\right \|^2_2 \right]
    \end{aligned}
\end{equation}
This final form is precisely the objective function presented in Eq.~\eqref{eqn:latent_conditionscorematching}. This proves that taking the expectation over the joint distribution $p(z^U_0, z^V)$ is mathematically equivalent to the ideal objective and gives a consistent training result, while being practically feasible to sample from.

\section{Details of the numerical solver}\label{sec: data_generation}
The data for the 2-D Navier--Stokes equation in Eq.~\eqref{2dNS} is generated using the pseudo-spectral method combined with the Crank-Nicolson scheme. 

\subsection{Pseudo-spectral solver}\label{ssec: pseudospectral}
We start with the initial condition $\omega(\mathrm{x}, t_0) \sim \mathcal{N}(0, 7^{3/2}(-\Delta + 49I)^{-5/2})$ with periodic boundary conditions, where $\Delta$ is a Laplace operator and $I$ is the identity operator. In Fourier space, we first obtain
\begin{equation}
    \hat{\omega}(k, t_0) = \mathcal{F}(\omega(\mathrm{x}, t_0)),
\end{equation}
where $\mathcal{F}$ denotes Fourier transformation, $\hat{\omega}$ denotes the Fourier coefficients of the vorticity field $\omega$, and $k = (k_x, k_y)$ represents wavenumbers, which are computed based on the grid size.

To approximate the convection and diffusion terms, we start by calculating positive Fourier multiplier associated with $-\nabla^2$, which is given by a constant
\begin{equation}
    C = 4\pi^2(k_x^2 + k_y^2).
\end{equation}
Then the stream function $\psi$ is obtained by solving the Poisson equation in Fourier space
\begin{equation}
    \hat{\psi} = \frac{\hat{\omega}}{C},
\end{equation}
with the velocity fields computed as
\begin{equation}\label{eqn: fourier_gradient}
    \begin{aligned}
        \hat{u} &= \left[\left(2\pi i k_y\right)\hat{\psi}, -\left(2\pi i k_x\right)\hat{\psi}\right].
    \end{aligned}
\end{equation}
The Fourier coefficients of the vorticity gradient, $\widehat{\nabla \omega} = \left[\left(2\pi i k_x\right)\hat{\omega},\left(2\pi i k_y\right)\hat{\omega}\right]$, are also calculated in Fourier space. Then, 
$\widehat{\nabla \omega}$ and $\hat{u}$ are converted back to the physical space via inverse Fourier transform to calculate the nonlinear convection term, which we denote as
\begin{equation}
    F(\mathrm{x}, t) = u(\mathrm{x}, t) \cdot \nabla \omega(\mathrm{x}, t)
\end{equation}

\subsection{Crank-Nicolson method}\label{ssec: cranknicolson}
The vorticity field is updated at each time step using the Crank-Nicolson scheme, which is implicit in time and second-order accurate:
\begin{equation}\label{eqn: CrankNicolson}
    \begin{aligned}
        \hat{\omega}(k, t_{n+1}) &= \frac{\hat{\omega}(k, t_{n}) - \Delta t \hat{F}(k, t_n) + \Delta t \hat{f}(k) - \frac{\Delta t}{2}\nu C \hat{\omega}(k, t_{n}) + \Delta t \beta \hat{\xi}_n}{1+\frac{\Delta t}{2} \nu C } \\
    \end{aligned}
\end{equation}
where $\nu = 10^{-3}$ is the viscosity, $\Delta t = 10^{-3}$ is the time step, $\hat{F}(k, t_n)$ is the Fourier transform of the nonlinear convection term, and $\hat{f}$ is the Fourier transform of the deterministic forcing term.

Recall that in Section~\ref{sec: numerical_results}, the closure term is defined as $H(\mathrm{x}, t) = - u(\mathrm{x}, t) \cdot \nabla\omega(\mathrm{x}, t) + \beta\xi$. Thus, when this closure is coupled to the solver, the above equation can be rewritten as
\begin{equation}
    \begin{aligned}
        \hat{\omega}(k, t_{n+1}) &= \frac{\hat{\omega}(k, t_{n}) - \frac{\Delta t}{2}\nu C\hat{\omega}(k, t_n) + \Delta t \hat{f}(k) + \Delta t\,\mathcal{F}[H](k, t_n)}{1+\frac{\Delta t}{2} \nu C}\\
        &\approx \frac{\hat{\omega}(k, t_{n}) - \frac{\Delta t}{2}\nu C\hat{\omega}(k, t_n) + \Delta t \hat{f}(k) + \Delta t\,\mathcal{F}[\hat{H}](k, t_n)}{1+\frac{\Delta t}{2} \nu C},
    \end{aligned}
\end{equation}
which represents using the learned closure sample $\hat{H}$ to simulate the 2-D Navier--Stokes system shown in Section~\ref{sec: surrogate}.

\section{Model architectures and training details}\label{sec:training_details}
Our latent space conditional diffusion framework in this work is built from three parts: two identical convolutional autoencoders -- one for the target fields $U$, and the other one for the conditional fields $V$, together with one conditional score-based diffusion model $s_\theta(\tau, z^U_\tau, z^V)$ built upon Fourier neural operators (FNOs).

The two autoencoders share the exact same architecture, where the encoder-decoder pair is a deep convolutional autoencoder built from residual and self‐attention blocks, mapping input fields of size $64\times64$ down to latent representations of size $16\times16$ and back. The encoders and decoders are symmetrically built from residual blocks—each combining GroupNorm, SiLU activations, and 3×3 convolutions—with strided downsampling in the encoder and matching upsampling in the decoder. A lightweight self‐attention module (four heads) is inserted at the bottleneck, projecting features to query/key/value vectors and then re‐projecting the attention output. 

The conditional score-based model $s_\theta(\tau, z^U_\tau, z^V)$ consists of two parallel FNO branches. In the first branch, the noisy target latent $z^U_\tau$ is augmented with sinusoidal Gaussian Fourier features of the diffusion time $\tau$~\cite{tancik2020fourier}, embedded via a small linear network, and concatenated with normalized spatial coordinates.  This allows the model to learn the dependency between noise levels $\tau$ and corresponding diffused states $z^U_\tau$. In the second branch, the conditional latent $z^V$ is likewise concatenated with grid coordinates.  Each branch processes its inputs through four Fourier layers—each performing a 1×1 grid-space convolution followed by a spectral-space convolution that applies learnable complex weights to the first 4 Fourier modes—interleaved with GELU activations.  The two branches are then merged by channel‐wise concatenation and refined through a 1×1 convolutional network to produce the final score function estimate.

For our 2-D NSE example, we simulate 100 trajectories with different initial conditions and extract the data spanning the physical time range of 20 s to 40 s, collecting 20,000 paired snapshots of resolved vorticity and corresponding closures $(\omega,H)$. For autoencoder pre-training in the two-phase pipeline, we split these samples into 80\% for training (18,000 snapshots), 10\% for validation, and 10\% for testing. Training is performed on NVIDIA GPUs using PyTorch with a batch size of 200, and optimization via Adam (base learning rate $l=10^{-3}$). We employ a Reduce-On-Plateau scheduler (halving the learning rate after 50 epochs with no improvement in validation loss) and monitor MSE on the held-out set for early stopping (patience of 100 epochs). Models are trained for up to 1000 epochs, with the best weights—those achieving the lowest validation loss—saved automatically, ensuring robust convergence before proceeding to latent-diffusion training. In the conventional two‐phase latent diffusion training, we encode all snapshots into $16\times16$ latent pairs $(z^\omega,z^H)$ and train the score-based diffusion model on 18,000 of these pairs for 500 epochs. We use Adam ($l = 10^{-3}$), a step‐decay schedule that halves the learning rate every 100 epochs, and a minibatch of 200.

For our end‐to‐end joint training paradigm, we optimize the autoencoder and diffusion components simultaneously using the multi‐objective loss in Eq.~\eqref{eqn: joint_loss_full}. The loss weights $\lambda_H=10$, $\lambda_\omega=0.1$, $\lambda_{\mathrm{score}}=0.1$, and $\lambda_{\mathrm{KL}}=0.01$ were selected via grid search on the validation set. Joint training operates on full‐resolution $(\omega, H)$ pairs at $64\times64$, incorporates a KL regularization on $z^H$ to prevent latent collapse, and otherwise follows the same optimizer and scheduling settings as above. The training and sampling algorithms are provided in the following appendix. This comprehensive protocol yields models that are both highly efficient in inference and robust in generative fidelity.

\section{Ablation studies on model architecture}\label{sec:ablation_studies}

\subsection{Ablation study on the necessity of non-local modeling}\label{sec: ablation_nonlocal}
To empirically validate our hypothesis that non-local modeling is a necessity for this physical system, we conducted an ablation study. We compare three distinct deterministic regression models, all tasked with learning the direct, physical-space mapping from the resolved state $\omega$ to the closure term $H$.

The primary difference between these models is their architectural capacity for non-local interactions. To ensure a fair comparison, all three models were designed with a comparable number of trainable parameters ($\sim$0.4M).
\begin{enumerate}
    \item FNO (Global): A deterministic Fourier Neural Operator, as described in the main text. Its spectral convolution is an inherently global operator, meaning the output at any point depends on the entire input domain.

    \item U-Net (Hierarchical-Local): A standard U-Net architecture. Its prediction at any point depends on a local neighborhood (its receptive field), which grows hierarchically through encoding and decoding.

    \item MLP (Pure-Local): A pixel-wise Multi-Layer Perceptron. The output at a point $(i, j)$ depends only on the input at that same point $(i, j)$ and its coordinates. This is a true local model.
\end{enumerate}

The models were trained to minimize the Mean Squared Error (MSE) between the predicted closure and the ground truth. The resulting test errors are presented in Table~\ref{tab:nonlocal_ablation}.

\begin{table}[H]
\centering
\caption{Comparison of deterministic closure models. The purely local MLP fails, while the global FNO performs best.}
\begin{adjustbox}{width=0.5\linewidth}
\begin{tabular}{@{}llcc@{}}
\toprule
\textbf{Model} & \textbf{Architecture Type} & \textbf{$D_\text{MSE}$} & \textbf{$D_\text{RE}$} \\
\midrule
\textbf{FNO} & \textbf{Global / Non-Local} & $\mathbf{4.10 \times 10^{-4}}$ & $\mathbf{0.0862}$ \\
U-Net & Hierarchical-Local & $5.09 \times 10^{-4}$ & $0.0955$ \\
MLP & Pure-Local (Pixel-wise) & $4.97 \times 10^{-2}$ & $2.6537$ \\
\bottomrule
\label{tab:nonlocal_ablation}
\end{tabular}
\end{adjustbox}
\end{table}

The results of this ablation study are unambiguous. The Pure-Local (MLP) model fails catastrophically to learn the mapping, with a relative error more than 27 times higher than the non-local models. This confirms that a local-in-space model is fundamentally insufficient for this problem. The Hierarchical-Local (U-Net) performs well, showing that its large receptive field is able to capture much of the necessary physics. However, the Global (FNO) model performs the best, achieving the lowest error in both metrics.

This study provides clear empirical evidence for two of our central claims:
\begin{enumerate}
    \item Non-local information is a prerequisite for accurately modeling the closure term in this system.
    \item An explicitly global operator (the FNO) is the most effective architecture for capturing these non-local dependencies.
\end{enumerate}

This study therefore validates the architectural choices made in our L-CDM framework, demonstrating that non-local operators are essential for accurately modeling the target system.

\subsection{Ablation study on the necessity of stochastic modeling}\label{sec: ablation_stochastic}

The first ablation study confirmed that a non-local model is necessary to capture the mean-field behavior. Our second study tests if a deterministic non-local model is sufficient, or if a stochastic framework is required.

To isolate this variable, we compare two models that share the same non-local FNO architecture:
\begin{enumerate}
\item P-FNO: The deterministic Physical-space FNO from the previous study (Table~\ref{tab:nonlocal_ablation}), trained with an MSE loss.
\item P-CDM: Our Physical-space Conditional Diffusion Model, trained with a score-matching loss to learn the full conditional distribution $p(H\mid\omega)$.
\end{enumerate}

The target closure $H(x,t)$ is an inherently stochastic field due to the $\beta\xi$ forcing term. The standard deviation of this field can be analytically calculated from the parameters of the $Q$-Wiener process used to generate the noise. Given the variance of the discrete-time noise $\mathrm{Var}(\xi^n) = \frac{\kappa}{L_1 L_2 \Delta t} \sum_{\mathbf{k}} q_{\mathbf{k}}$, we use our numerical setup parameters (amplitude $\beta=5\times 10^{-5}$, time step $\Delta t=10^{-3}$, variance inflation $\kappa=10$, domain $L_1=L_2=1$, and $\sum_{\mathbf{k}} q_{\mathbf{k}} \approx 16.0$) to find the standard deviation of the closure component:

\begin{equation}
    \begin{aligned} 
        \mathrm{Std}(\beta\xi^n) &= \beta \cdot \mathrm{Std}(\xi^n) = \beta \cdot \sqrt{\mathrm{Var}(\xi^n)} \\
        &= \beta \cdot \sqrt{\frac{\kappa}{L_1 L_2 \Delta t} \sum_{\mathbf{k}\in\mathcal{K}} q_{\mathbf{k}}} \\ 
        &\approx (5\times 10^{-5}) \cdot \sqrt{\frac{10}{1 \cdot 1 \cdot 10^{-3}} \times 16.0} = 0.02. 
    \end{aligned}
\end{equation}

\begin{table}[H]
\centering
\caption{Comparison of deterministic (P-FNO) vs. stochastic (P-CDM) non-local models. Both models capture the mean-field behavior, but only the stochastic P-CDM can reproduce the system's physical variance (Theoretically $\approx 0.02$).}
\begin{adjustbox}{width=0.5\linewidth}
\label{tab:stochastic_ablation}
\begin{tabular}{@{}llcc@{}}
\toprule
\textbf{Model} &\textbf{Model Type} & \textbf{$D_\text{RE}^{\text{ens}}$} & \textbf{Field Std.} \\
\midrule
P-FNO & Deterministic & 8.620e-02 & 0.0 (by definition) \\
\textbf{P-CDM} & \textbf{Stochastic} & \textbf{6.851e-02} & \textbf{1.889e-02} \\
\bottomrule
\end{tabular}
\end{adjustbox}
\end{table}

While both models are proficient at capturing the mean behavior (with comparable relative errors), the deterministic P-FNO is, by definition, incapable of reproducing the physical variance, reporting a standard deviation of 0.0. Our stochastic P-CDM, in contrast, generates an ensemble with a standard deviation of 1.889e-02, in excellent agreement with the theoretical value of 0.02 as shown in Table~\ref{tab:stochastic_ablation}. Thus, a stochastic generative framework (like our P-CDM or Joint L-CDM) is essential to capture the full, physically correct statistical distribution and uncertainty.

\section{Learning stochastic closures for under-resolved Large Eddy simulations}
\label{sec:LES_closure}

\subsection{Problem formulation and closure definition}
Here we consider the two-dimensional incompressible Navier--Stokes equations in
vorticity form, as a true system with deterministic governing equations:
\begin{equation}
\label{eq:appendix_eq1}
\frac{\partial \omega}{\partial t} + u \cdot \nabla \omega
= \nu \nabla^2 \omega + f(\mathbf{x}),
\end{equation}
with a square periodic domain of side length $L=1$ and viscosity coefficient $\nu = 10^{-3}$. The deterministic forcing used in the simulations was
\begin{equation}
f(\mathbf{x}) = 0.1\left(\sin(2\pi(x+y)) + \cos(2\pi(x+y))\right).
\end{equation}
The initial vorticity field is sampled from a Gaussian random field
\begin{equation}
\omega_0 \sim \mathcal{N}\left(0,\, 7^{3/2}(-\Delta + 49I)^{-5/2}\right).
\end{equation}

High-fidelity trajectories are generated by solving Eq.~\eqref{eq:appendix_eq1} on a uniform grid with resolution $N_\mathrm{hr} = 2048$ in each spatial direction. To obtain the resolved (coarse-grained) variables, we apply a spatial low-pass filter $\bar{(\cdot)}=\mathcal{G}_{\Delta}(\cdot)$ yielding
\begin{equation}
\frac{\partial \bar{\omega}}{\partial t} + \overline{u \cdot \nabla \omega}
= \nu \nabla^2 \bar{\omega} + \bar{f}(\mathbf{x}),
\end{equation}
which can be rewritten as
\begin{equation}
\frac{\partial \bar{\omega}}{\partial t} + \bar{u} \cdot \nabla \bar{\omega}
= \nu \nabla^2 \bar{\omega} + \bar{f}(\mathbf{x}) + U,
\end{equation}
with the exact closure term defined as:
\begin{equation}
U=\bar{u} \cdot \nabla \bar{\omega}-\overline{u \cdot \nabla \omega}.
\end{equation}

The filtering operation is implemented in Fourier space using a Gaussian kernel
\begin{equation}
\widehat{\mathcal{G}_{\Delta}}(k)
=
\exp\!\left(-\frac{\Delta^2 |k|^2}{24}\right),
\end{equation}
which, for the present domain and grid,
corresponds to a filter width $\Delta=0.25$. After filtering, the fields are spectrally truncated to a coarse resolution of $N_\mathrm{cr} = 64$, corresponding to a downsampling factor of $32$. The filtered vorticity $\bar{\omega}$, filtered forcing $\bar{f}$, and exact closure term $U$ are all computed from the high-fidelity trajectories using this consistent filtering and truncation procedure.

It is worth noting that distinct high-fidelity states from the true system may correspond to the same filtered field of $\bar{\omega}$ while leading to different fields of $U$, and the variability can be characterized by $p(U\mid\bar{\omega})$.

\subsection{A-priori conditional generation results}
We compare the physical-space
conditional diffusion model (P-CDM), the conventional two-phase latent conditional diffusion
model (Two-Phase L-CDM), and the jointly trained latent conditional diffusion model (Joint
L-CDM). For all qualitative visualizations in this numerical example, we report ensemble-mean predictions estimated from $1000$ conditional samples.

\begin{table}[H]
\centering
\caption{
Performance comparison of P-CDM, Two-Phase L-CDM, and Joint L-CDM on learning the stochastic closures for under-resolved LES. The reported metrics include reconstruction errors of the closure term,
generation errors in the latent space, generation errors in the physical space, and the
computational cost of drawing $1000$ conditional samples.
}
\begin{adjustbox}{width=0.9\linewidth}
\begin{tabular}{@{}cccccccc@{}}
\toprule
\multirow{3}{*}{\textbf{Model}}         & \multicolumn{2}{c}{\textbf{Recon Err of $U$}} & \multicolumn{2}{c}{\begin{tabular}[c]{@{}c@{}}\textbf{Latent Space}\\ \textbf{Generation}\end{tabular}} & \multicolumn{2}{c}{\begin{tabular}[c]{@{}c@{}}\textbf{Physical Space}\\ \textbf{Generation}\end{tabular}} & \multirow{2}{*}{\begin{tabular}[c]{@{}c@{}}\textbf{Cost}\\ \textbf{(s/1000)}\end{tabular}} \\ \cmidrule(lr){2-7}
                & $D_\text{MSE}$    & $D_\text{RE}$    & $D_\text{MSE}$                             & $D_\text{RE}$                            & $D_\text{MSE}$                              & $D_\text{RE}$                             &                                                                          \\ \midrule
P-CDM           & -                 & -                & -                                          & -                                        & 4.542e-07                                   & 1.182e-02                                 & 2.86                                                                     \\
Two-Phase L-CDM & 2.907e-06         & 6.965e-03        & 5.422e-04                                  & 6.142e-02                                & 6.853e-06                                   & 4.593e-02                                 & 0.43                                                                     \\
Joint L-CDM     & 4.026e-06         & 8.711e-03        & 1.456e-05                                 & 3.936e-03                               & 9.101e-08                                   & 8.342e-03                                & 0.41                                                                     \\ \bottomrule
\end{tabular}
\end{adjustbox}
\label{tab:LES_Comparison}
\end{table}

\begin{figure}[!htbp]
    \centering
    \includegraphics[width=\linewidth]{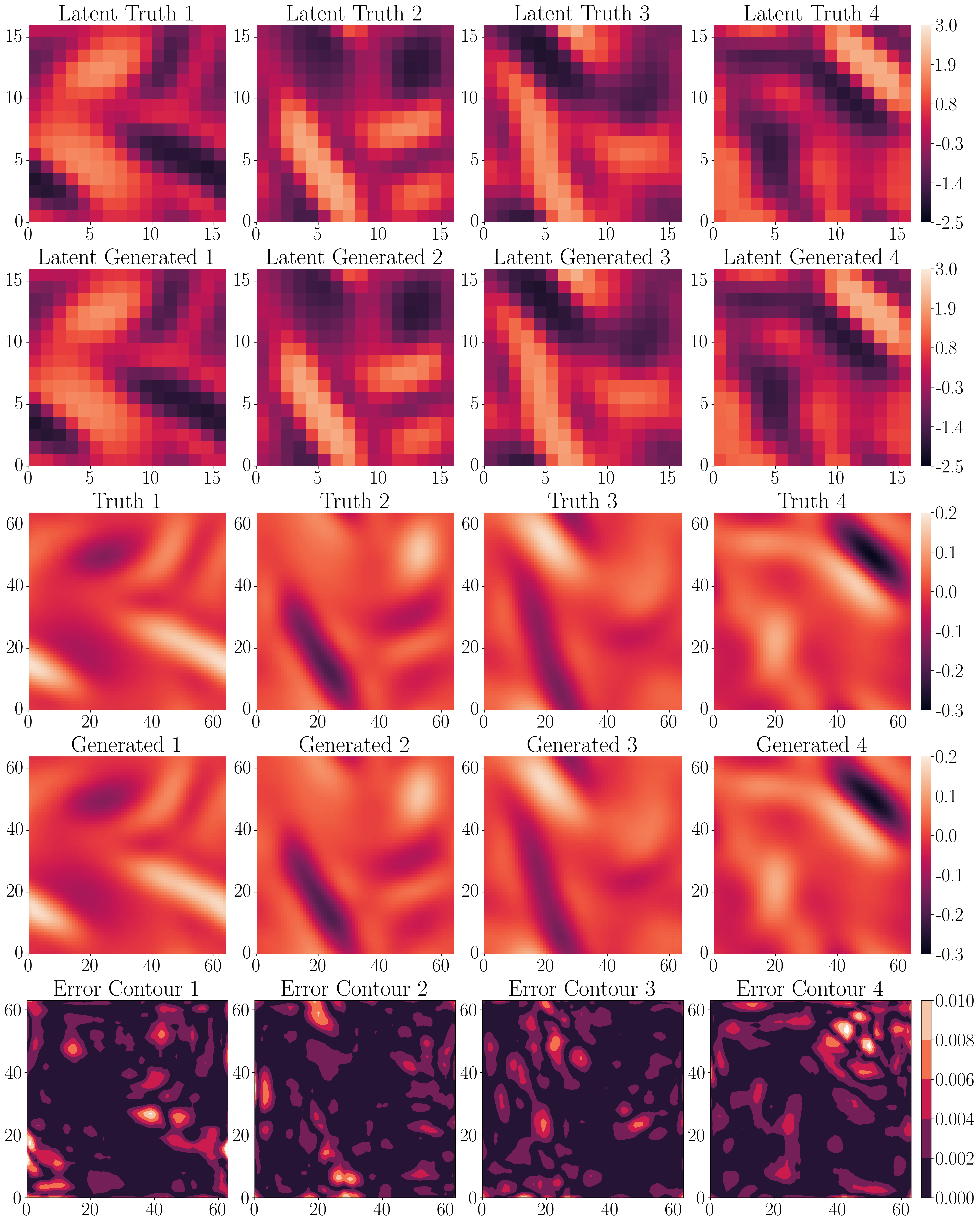}
    \caption{Conditional generation results of the conventional Two-Phase L-CDM for the LES closure task.
    \textbf{First row:} encoded ground-truth latent representations $z^U$.
    \textbf{Second row:} ensemble-mean latent predictions $\mathbb{E}_{\mathrm{MC}}[\hat{z}^U \mid z^{\bar{\omega}}]$ estimated from $1000$ conditional samples.
    \textbf{Third row:} ground-truth closure terms $U$ computed directly from filtered high-fidelity data.
    \textbf{Fourth row:} ensemble-mean decoded closure fields $\mathbb{E}_{\mathrm{MC}}[\hat{U} \mid \bar{\omega}]$ estimated from $1000$ conditional samples.
    \textbf{Fifth row:} absolute error fields between the exact closure and the ensemble-mean generated field.
    Different columns correspond to different test snapshots.}
    \label{fig:LESWithoutJoint}
\end{figure}

\begin{figure}[!htbp]
    \centering
    \includegraphics[width=\linewidth]{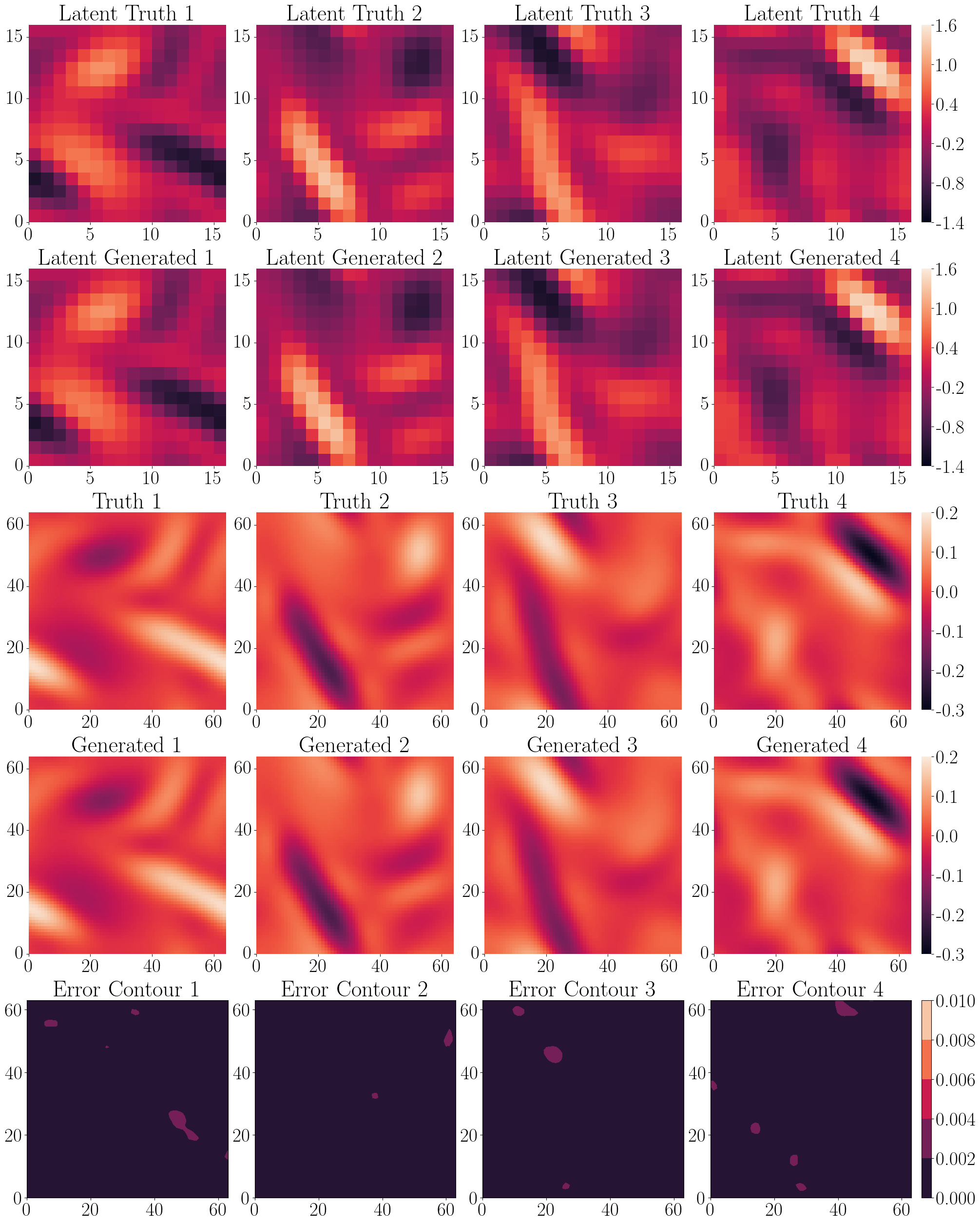}
    \caption{Conditional generation results of the proposed Joint L-CDM for the LES closure task.
    \textbf{First row:} encoded ground-truth latent representations $z^U$.
    \textbf{Second row:} ensemble-mean latent predictions $\mathbb{E}_{\mathrm{MC}}[\hat{z}^U \mid z^{\bar{\omega}}]$ estimated from $1000$ conditional samples.
    \textbf{Third row:} ground-truth closure terms $U$ computed directly from filtered high-fidelity data.
    \textbf{Fourth row:} ensemble-mean decoded closure fields $\mathbb{E}_{\mathrm{MC}}[\hat{U} \mid \bar{\omega}]$ estimated from $1000$ conditional samples.
    \textbf{Fifth row:} absolute error fields between the exact closure and the ensemble-mean generated field.
    Different columns correspond to different test snapshots.}
    \label{fig:LESWithJoint}
\end{figure}

Table~\ref{tab:LES_Comparison} summarizes the performance of various methods on learning the stochastic closures for under-resolved LES, which leads to the same conclusion as the numerical example studied in Section~\ref{sec: numerical_results}. More specifically, 
P-CDM remains a strong physical-space baseline but is relatively expensive. On the other hand, Two-Phase L-CDM
greatly reduces the sampling cost and yields good reconstruction accuracy, yet performs much
worse on the generative task. In contrast, Joint L-CDM produces the best overall balance:
it preserves the computational advantage of latent-space sampling while substantially improving
both latent-space and physical-space generation accuracy. More detailed results of the generated closure terms are presented in Figs.~\ref{fig:LESWithoutJoint} and~\ref{fig:LESWithJoint}.

\subsection{A-posteriori LES simulations with trained stochastic closures}

We next assess the \textit{a posteriori} performance of the learned stochastic closures by coupling
them with the LES solver and evolving the resolved dynamics forward in time. The goal is to
determine whether the learned conditional closure can recover the filtered results of the true system. For all results reported in this section, the applied closure
correction is based on the ensemble mean estimated from $1000$ independently generated conditional samples.

\begin{figure}[!htbp]
    \centering
    \includegraphics[width=\linewidth]{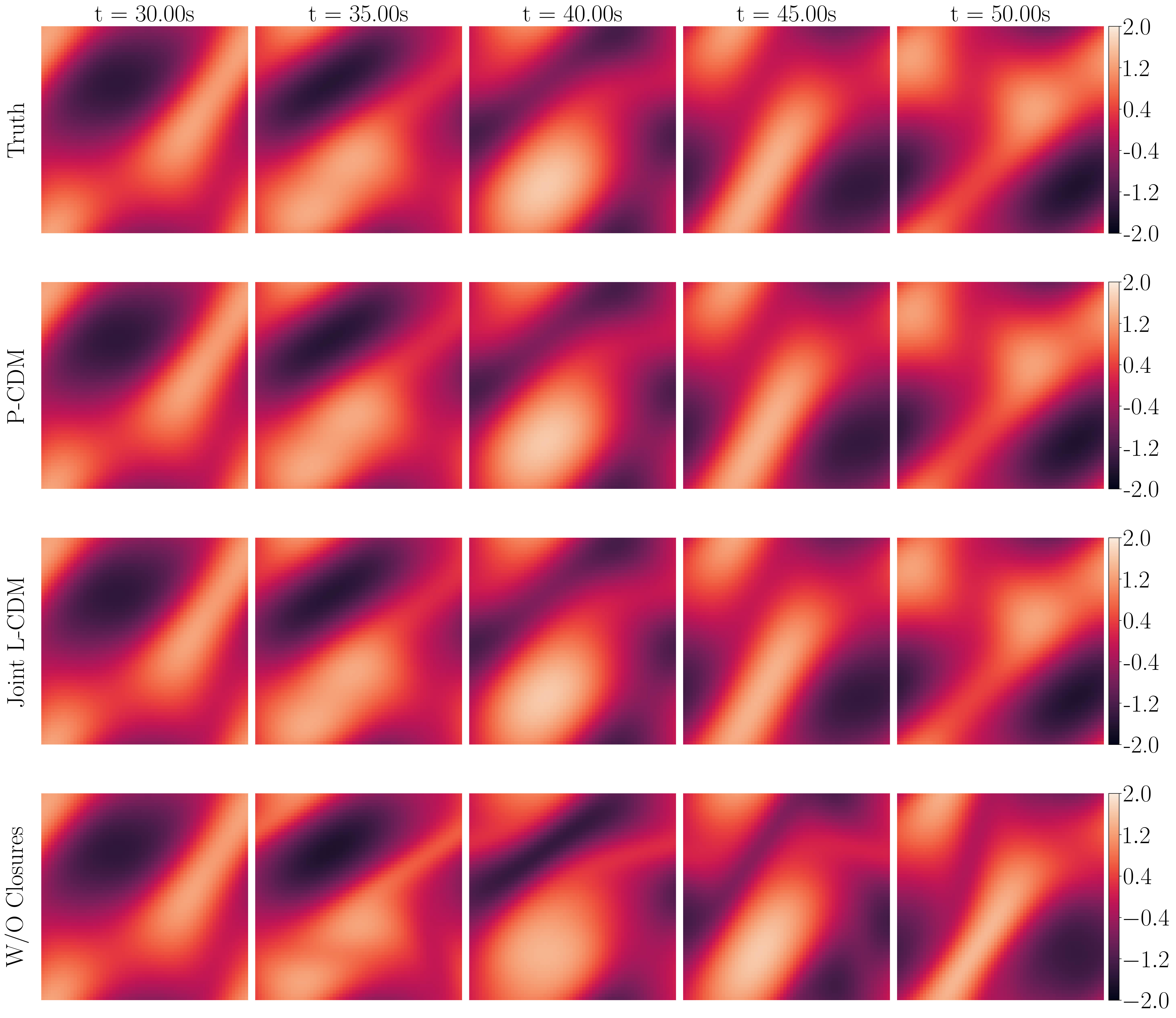}
    \caption{Temporal evolution of the resolved vorticity field over a 20-second simulation window. \textbf{First row:} filtered high-fidelity ground truth. \textbf{Second row:} LES simulation corrected using the P-CDM ensemble-mean closure estimated from 1000 conditional samples. \textbf{Third row:} LES simulation corrected using the Joint L-CDM ensemble-mean closure estimated from 1000 conditional samples. \textbf{Fourth row:} LES simulation without closure correction.}
    \label{fig: surrogate_LES}
\end{figure}

Figure~\ref{fig: surrogate_LES} compares the resulting vorticity fields over 20 time units, starting from $t=30$. Both P-CDM and Joint L-CDM recover the main structures of the ground truth. In contrast, the simulation without closure progressively departs from the ground truth, confirming that the closure term is important to ensure a good match with the true system. Table~\ref{tab:les_simulation_metrics} further presents the time costs and the quantitative performance metrics. The no-closure model is indeed
the least expensive, but it is also substantially less accurate: its relative vorticity error grows
from $1.94\times 10^{-1}$ at $t=35$ to $6.12\times 10^{-1}$ at $t=50$, whereas both stochastic
closure models remain at the level of $O(10^{-2})$. Among the two stochastic models, Joint L-CDM reduces the total ensemble simulation cost from $8701.23$ s to $1198.65$ s,
corresponding to a speedup of about $7.3\times$. Although the time cost of joint L-CDM is noticeably higher than the model without any correction for this canonical 2-D example, it is expected that the computational cost of under-resolved LES with the proposed stochastic closures is much less than the high-fidelity simulations (e.g., fully resolved LES or direct numerical simulation) for most 3-D turbulent flow problems in real-world applications.

\begin{table}[!htbp]
\centering
\caption{Simulation performance over a 20-second integration. Reported are ensemble mean prediction errors based on averages over 1000 independently simulated trajectories. The reported time cost is the total wall-clock time required to generate and average 1000 samples for the ensemble simulation.}
\begin{adjustbox}{width=\linewidth}
\begin{tabular}{lcccccccc}
\toprule
\multirow{2}{*}{\textbf{Model}} 
& \multirow{2}{*}{\textbf{Cost (s)}} 
& \multirow{2}{*}{\textbf{Metric}} 
& \multirow{2}{*}{\textbf{Closure term}} 
& \multicolumn{5}{c}{\textbf{Vorticity field error at time}} \\
\cmidrule(lr){5-9}
& & & & \textbf{t=30} & \textbf{t=35} & \textbf{t=40} & \textbf{t=45} & \textbf{t=50} \\
\midrule

\multirow{2}{*}{W/O Closures} 
& \multirow{2}{*}{16.33} 
& $D_\text{RE}$ & -- 
& 0 & 1.94e-01 & 4.10e-01 & 4.42e-01 & 6.12e-01 \\
& 
& $D_\text{MSE}$ & -- 
& 0 & 2.18e-02 & 9.63e-02 & 1.07e-01 & 2.04e-01 \\
\midrule

\multirow{2}{*}{P-CDM} 
& \multirow{2}{*}{8701.23} 
& $D_\text{RE}$ & 1.18e-02 
& 0 & 1.07e-02 & 1.43e-02 & 1.90e-02 & 2.08e-02 \\
& 
& $D_\text{MSE}$ & 4.54e-07 
& 0 & 6.57e-05 & 1.17e-04 & 1.98e-04 & 2.36e-04 \\
\midrule

\multirow{2}{*}{Joint L-CDM} 
& \multirow{2}{*}{1198.65} 
& $D_\text{RE}$ & 8.34e-03 
& 0 & 4.00e-03 & 3.70e-03 & 7.90e-03 & 1.17e-02 \\
& 
& $D_\text{MSE}$ & 9.10e-08 
& 0 & 9.28e-06 & 7.74e-06 & 3.42e-05 & 7.45e-05 \\
\bottomrule
\end{tabular}
\end{adjustbox}
\label{tab:les_simulation_metrics}
\end{table}

Figure~\ref{fig: TKE_LES} further compares the kinetic energy spectra at $t=30$, $t=40$, and
$t=50$. Since all models share the same initial condition, the spectra coincide at $t=30$. As
the simulations evolve, the no-closure model departs rapidly from the reference spectrum,
whereas both stochastic closure models remain much closer, especially in the low- and 
intermediate-wavenumber ranges that the under-resolved LES without a closure model fails to capture well.

\begin{figure}[!htbp]
    \centering
    \includegraphics[width=\linewidth]{TKE_Closure_3050_LSE.png}
    \caption{Comparison of kinetic energy spectra at three representative times, $t=30$, $t=40$, and $t=50$, for the filtered high-fidelity ground truth, the P-CDM, the Joint L-CDM, and the simulation without closures (W/O Closures). The spectra are plotted in log--log scale against wave number $k$.}
    \label{fig: TKE_LES}
\end{figure}
\clearpage
\bibliographystyle{model1-num-names}
\bibliography{references}
\end{document}